\documentclass{article}

    \PassOptionsToPackage{numbers, compress}{natbib}
\usepackage[final]{neurips_2025}

\usepackage[utf8]{inputenc} % allow utf-8 input
\usepackage[T1]{fontenc}    % use 8-bit T1 fonts
\usepackage{hyperref}       % hyperlinks
\usepackage{url}            % simple URL typesetting
\usepackage{booktabs}       % professional-quality tables
\usepackage{amsfonts}       % blackboard math symbols
\usepackage{nicefrac}       % compact symbols for 1/2, etc.
\usepackage{microtype}      % microtypography
\usepackage{xcolor}         % colors

\usepackage[table]{xcolor} % 添加颜色支持
\usepackage{multirow}
\usepackage{graphicx}
\usepackage{amsmath}
\usepackage{booktabs} % 用于\toprule等命令
\usepackage{wrapfig}
\usepackage{amsmath}
\usepackage{colortbl}   % 支持单元格背景颜色
\usepackage{booktabs}   % 提供更漂亮的横线
\usepackage{multirow}   % 支持单元格合并
\usepackage{pifont}     % 支持特殊符号，如 \ding{55}
\usepackage{caption}    % 美化表格标题
\usepackage{xcolor}     % 支持颜色定义
\usepackage{wrapfig} % 在导言区添加
\usepackage{tabularx}
\usepackage{caption}
\usepackage{float} % 添加float包
\usepackage{placeins} % 添加placeins包

\renewcommand{\thefootnote}{\fnsymbol{footnote}}

\title{MUVR: A Multi-Modal Untrimmed Video Retrieval Benchmark with Multi-Level Visual Correspondence}
\begin{document}

\author{
  Yue Feng\textsuperscript{1},
  Jinwei Hu\textsuperscript{1},
  Qijia Lu\textsuperscript{1},
  Jiawei Niu\textsuperscript{1},
  Li Tan\textsuperscript{1},
  Shuo Yuan\textsuperscript{1},  \\
  \textbf{Ziyi Yan\textsuperscript{1},}
  \textbf{Yizhen Jia\textsuperscript{1},}
  \textbf{Qingzhi He\textsuperscript{1},}
  \textbf{Shiping Ge\textsuperscript{2},}  \\
  \textbf{Ethan Q. Chen\textsuperscript{3},}
  \textbf{Wentong Li\textsuperscript{1\dag},}
  \textbf{Limin Wang\textsuperscript{2},}
  \textbf{Jie Qin\textsuperscript{1\dag}} \\
  % \smallskip
  \textsuperscript{1}MoE Key Laboratory of Brain-Machine Intelligence Technology, \\
  College of Artificial Intelligence, Nanjing University of Aeronautics and Astronautics \\
  \textsuperscript{2}Nanjing University
  \textsuperscript{3}The Hong Kong Polytechnic University
  % \smallskip
  % \texttt{fengyue@nuaa.edu.cn}, \texttt{hujinwei0310@nuaa.edu.cn}, \texttt{qinjiebuaa@gmail.com}
}

% 添加脚注
\renewcommand{\thefootnote}{\dag} % 将脚注符号设置为 †
\footnotetext[1]{Corresponding authors.}

\maketitle
% \footnotetext[†]{These authors contributed equally to this work.}
% 对比相关benchmark
% 评测需要完善合理
% 这个领域的数据方法有哪些
% 提出新数据和评测的动机是否强烈，即提出的DB是对相关领域的痛点（例如缺少哪方面数据内容/更多标签/做数据的工具等）很好的缓解或解决，动机要先描述该领域卡在了某方面数据上或领域发展太快缺少公平对比的榜单、原本的评价指标无法突出现在算法的差异等
% 明确工作的贡献123…做数据的一般从新数据规模、质量、新标注（Motion-X）or 数据内容解决了某些问题（Human-Art）、都要对数据的统计分析与信息挖掘及来、并对对提出数据在哪些任务上验证了其有效性。做benchmark需要从现存该领域的方法（往往是整理近期涌现的工作）、到不同数据和指标上进行统一评测（展现工作量），评测的角度、覆盖范围及结论是否有新的信息和启发、是否给社区带来新的思考决定了这篇文章的贡献程度。benchmark也可以是基于大规模数据和评测带来方法上的探索，例如利用大规模数据（需要侧重对数据的分析）更好地scale up、对方法的思考（SMPLer-X）
% 以数据集为例，文章思路：过去该领域发展遇到了123困难，目前的数据存在456问题->本文提出了对应的n个方面来解决这456问题->具体这n方面数据上怎么做的、有什么挑战、怎么解决的->分析、统计、对比、展示提出的数据->实验验证数据的有效性（要跟现有数据集做对比，例如用同样的方法，不同数据训练，查看新数据带来的收益），呼应是否缓解了该领域的123问题、从缓解的程度来反映文章贡献
% 收集的数据都需要列出数据来源、对应的license和版权问题以及潜在的bias，分发的时候没授权的部分只能给原始链接
% 官方给的一个关键标准是可访问性：数据集应该是可用和可访问的，任何所需的代码都应该是开源的，在补充材料要尽可能详细地介绍如何收集和组织数据的，其中包含哪些信息，如何以合乎道德和负责任的方式使用这些数据，以及如何提供和维护这些数据

% \begin{figure*}[h!]
%     \centering
%     \includegraphics[width=1\linewidth]{figs/fig1_partition.png}
%     \caption{
%     % Comparison between traditional fine-grained video retrieval and our fine-grained web video retrieval. Examples of query videos (top) and matches (bottom) under various retrieval-level.
%     Comparison of traditional fine-grained video retrieval \textit{vs.} our fine-grained web video retrieval: Illustrated above (ours) are sample query videos (top) alongside their corresponding matches (bottom) under different retrieval tiers.
%     }
%     \label{fig1}
% \end{figure*}

\begin{abstract}

We propose the Multi-modal Untrimmed Video Retrieval task, along with a new benchmark (MUVR) to advance video retrieval for long-video platforms. MUVR aims to retrieve untrimmed videos containing relevant segments using multi-modal queries. It has the following features: \textbf{1) Practical retrieval paradigm:} MUVR supports video-centric multi-modal queries, expressing fine-grained retrieval needs through long text descriptions, video tag prompts, and mask prompts. It adopts a one-to-many retrieval paradigm and focuses on untrimmed videos, tailored for long-video platform applications. \textbf{2) Multi-level visual correspondence:} To cover common video categories (e.g., news, travel, dance) and precisely define retrieval matching criteria, we construct multi-level visual correspondence based on core video content (e.g., news events, travel locations, dance moves) which users are interested in and want to retrieve. It covers six levels: copy, event, scene, instance, action, and others. \textbf{3) Comprehensive evaluation criteria:} We develop 3 versions of MUVR (i.e., Base, Filter, QA). MUVR-Base/Filter evaluates retrieval models, while MUVR-QA assesses MLLMs in a question-answering format. We also propose a Reranking Score to evaluate the reranking ability of MLLMs. MUVR consists of 53K untrimmed videos from the video platform Bilibili, with 1,050 multi-modal queries and 84K matches. Extensive evaluations of 3 state-of-the-art video retrieval models, 6 image-based VLMs, and 10 MLLMs are conducted. MUVR reveals the limitations of retrieval methods in processing untrimmed videos and multi-modal queries, as well as MLLMs in multi-video understanding and reranking. Our code and benchmark is available at~\url{https://github.com/debby-0527/MUVR}.

\end{abstract}

\setlength{\textfloatsep}{16pt} % 控制浮动体与正文间距（默认 20pt）
\setlength{\floatsep}{12pt}     % 控制双图之间的垂直间距（默认 12pt）
\setlength{\intextsep}{10pt}   % 控制浮动体在文本中的上下间距（默认 10pt）

% \setlength{\parskip}{5pt} % 取消段落间距

% 减少行间公式的上下间距
% \setlength{\abovedisplayskip}{5pt}   % 公式上方间距
% \setlength{\belowdisplayskip}{5pt}   % 公式下方间距
% \setlength{\abovedisplayshortskip}{1pt} % 短公式上方间距
% \setlength{\belowdisplayshortskip}{1pt} % 短公式下方间距

% \begin{figure*}[!t]
%     \centering
%     \includegraphics[width=1\linewidth]{figs/fig2_query.png}
%     \caption{Examples of Video Query, Text Query, and our Fine-Grained Composed Query. Video Query and Text Description are used in MUVR-Base. Tag Prompt is used in MUVR-Filter.}
%     \label{fig2}
% \end{figure*}

\section{Introduction}
\label{sec:intro}
The rapid growth of video platforms like YouTube, TikTok, and Bilibili has led to millions of videos being uploaded daily. Efficient retrieval of relevant videos from a video library is crucial for recommendation systems, content search~\cite{xu2024fine,Jia_Quan_Feng_Chen_Qin_2025,su2024maca,10817642}, and video understanding applications~\cite{ji2023multispectral,ge2025implicitlocationcaptionalignmentcomplementary,RefineTAD,10.1145/3581783.3612506}. It naturally raises three critical research questions: 1) What video retrieval paradigm best aligns with real-world applications? 2) How can benchmarks comprehensively cover diverse video categories across a video platform? 3) What are the performance limitations of current retrieval models? Current researches have certain limitations on the retrieval paradigm, video diversity, or evaluation criteria. To address these questions, we propose the Multi-modal Untrimmed Video Retrieval task and its corresponding benchmark (MUVR), which features the following three key characteristics:

\begin{table}[!t]
\caption{Comparison of existing video retrieval tasks and our MUVR. O2M: One-to-Many retrieval. UV: Untrimmed Video. T: Text. V: Video. Multi-partition: MUVR comprises five partitions (i.e., news, region, instance, dance, and others), each of which includes several categories of videos.}
\centering
% \small
\footnotesize
\setlength{\tabcolsep}{0.9mm}
\begin{tabular}{l|ccrrr}
\toprule
Video Retrieval Tasks & O2M & UV & Query & Category & Matching Criterion \\
\midrule
Text-to-Video Retrieval\cite{FCA-Net} & \ding{55} & \ding{55} & T & mixed & global semantic match \\
Composed Video Retrieval \cite{covr} & \ding{55} & \ding{55} & T, V & mixed & video modification\\
Partially Relevant Video Retrieval \cite{partiallyrelevantvideoretrieval} & \ding{55} & \checkmark & T & mixed & partial semantic match\\
Near-duplicate Video Retrieval\cite{UQ_VIDEO} & \checkmark & \checkmark & V & mixed & almost identical segment \\
Fine-grained Video Retrieval\cite{fivr} & \checkmark & \checkmark & V & news & same incident segment \\
% EVVE\cite{evve} & 13 & 620 & 1,252 & 102,375$^{\dagger}$ & YT \\
% FIVR\cite{fivr} & 100 & 100 & 12,300 & 225,960$^{\dagger}$ & YT \\
\midrule
\textbf{MUVR (Ours)} & \checkmark & \checkmark & T, V & multi-partition & multi-level visual correspondence \\
\bottomrule
\end{tabular}
\label{tab:vr_tasks_comparison}
\end{table}

\textbf{Practical Retrieval Paradigm.} 
% 我们在Table 1 展示了现存的视频检索任务的检索范式。由于用户习惯使用文本表达检索需求，很多视频检索研究关注文本query并根据全局/局部语义匹配来检索视频。然而，仅靠文本query难以描述细节视觉信息，往往导致检索范围过大或者使得文本query更繁琐。相比之下，仅靠视频query会引入无关的视觉信息，导致检索失败。一些任务利用完整的query视觉信息来检索，但这限制它们只适用于almost identical segment的检索或者特殊的视频类型场景。因此，合理的做法是用video query 表达文本难以描述的视觉细节，用文本query聚焦于关键视觉内容。考虑到视频平台中的视频有大量文本难以完整且准确描述的内容（例如新闻事件、不认识的特殊产品、流行元素），我们将视频query作为主导，文本query作为辅助。此外，为了筛选检索得到的视频，我们提出了一种操作简单的tag prompt，用户仅需给出希望视频具备/不具备的特征即可进行更精细的检索。这种方式类似于\cite{covr}，但是我们对更有挑战性的untrimmed video library进行一对多检索并且有大量人工标注的困难tag。为了进一步增强文本query对视频的细粒度指代表示，我们提出mask prompt来引导检索模型关注视频关键区域。
We present the retrieval paradigms of existing video retrieval tasks in Table \ref{tab:vr_tasks_comparison}. Since users are accustomed to expressing retrieval needs through text, many video retrieval studies focus on text queries and retrieve videos based on global/partial semantic matching \cite{FCA-Net, FrameFusionMoE,learningtexttovideoretrievalimage,TC-MGC,partiallyrelevantvideoretrieval, DLDKD}. However, text queries alone struggle to describe detailed visual information, often resulting in overly broad retrieval ranges or cumbersome text queries. In contrast, relying solely on video queries introduces irrelevant visual information, leading to retrieval failures. Some tasks utilize complete visual query information for retrieval, but this limits them to almost identical segment retrieval or specialized video categories like news. Therefore, a reasonable approach is to use video queries to express visual details that are difficult to describe in text, while using text queries to focus on key visual content. Considering that videos on platforms contain substantial content that is difficult to describe completely and accurately through text (e.g., news events, unfamiliar special products, and popular elements), we employ video queries as the dominant approach with text descriptions as the auxiliary. Additionally, to filter retrieved videos, we propose an easy-to-use tag prompt where users only need to specify desired/undesired video features for more refined retrieval. This approach resembles \cite{covr}, but we perform one-to-many retrieval on a more challenging untrimmed video library with numerous manually annotated difficult tags. To further enhance the text descriptions' fine-grained reference representation to video queries, we propose a mask prompt to guide the retrieval model's attention to key video regions.
% Existing FVR benchmarks, such as EVVE \cite{evve} and FIVR \cite{fivr}, rely on two coarse types of retrieval: frame-level and event-level. As shown in Figure~\ref{fig1}, frame-level retrieval focuses solely on duplicate frames matching with low-level pixels, while event-level retrieval is limited to time-sensitive news videos with spatio-temporal intersection. 
% These limitations make existing FVR methods \cite{dns, s2vs} unsuitable for most web videos, which are rich in semantics and time-insensitive. 
% To address this, \textcolor{red}{we decouple event-level retrieval into three time-insensitive levels}: region-level, instance-level, and action-level. Based on the expanded retrieval levels, we organize web videos into five partitions (\textit{i.e.}, news, region, instance, dance, entertainment) to build our \textcolor{red}{multi-partition fine-grained web video retrieval benchmark MUVR}.

\textbf{Diverse Video Categories and Multi-Level Visual Correspondence.}
% 如表一所示，不同的视频检索任务基于不同的检索匹配标准。基于语义的匹配适合能以文本概括的简单视频，在现实应用中受到限制。\cite{covr}旨在根据视频query和变化文本，检索经过变化后的视频，局限于一对一检索。基于视频query的任务仅适用于News类型视频或者片段重合的视频检索。为了cover more video categories and 准确构建更普适的检索匹配标准，我们提出multi-level visual correspondence. 具体而言，用户感兴趣并希望检索的视频内容通常包括底层的帧、中层语义的scene, instance, action以及上层语义的event和others。这对应六个levels of Visual Correspondence（i.e., copy, scene, instance, action, event and others）。虽然各种类型的视频都存在这些内容，用户可以从任意level出发检索相关视频。但是不同类型的视频有不同的用户希望检索的显著内容。因此，我们设计了五个分区（i.e., news, region, instance, dance, others）来划分benchmark并覆盖各种视频类型。例如，instance分区以动物、商品、食品等类型视频为主，专门用于instance-level retrieval。请参考第三节以获取更多细节。
As shown in Table \ref{tab:vr_tasks_comparison}, different video retrieval tasks are based on different retrieval matching criteria. Semantic-based matching is suitable for simple videos that can be summarized in text, but is limited in real-world applications. Ventura, etc. \cite{covr} aims to retrieve modified videos based on video queries and text modifications, but it is restricted to one-to-one retrieval. Video query-based tasks are only applicable to news-type videos or videos with overlapping segments. To cover more video categories and accurately establish a more universal retrieval matching criterion, we propose multi-level visual correspondence. Specifically, the video content that users are interested in and wish to retrieve typically includes low-level frames, mid-level semantics such as scene, instance, and action, as well as high-level semantics such as event and others. This corresponds to six levels of visual correspondence (i.e., copy, scene, instance, action, event, and others). Although these elements exist in various categories of videos and users can retrieve relevant videos from any level, different types of videos have distinct salient content that users prefer to retrieve. Therefore, we design five partitions (i.e., news, region, instance, dance, others) to categorize our MUVR benchmark and cover diverse video categories. For example, the instance partition primarily includes videos of pets, goods, etc., specifically designed for instance-level retrieval. Please refer to Section \ref{sec:MUVR Benchmark} for more details.

% As shown in Figure \ref{fig2}, current video retrieval tasks use either video \cite{CC_WEB_VIDEO, evve, vcdb, fivr, UQ_VIDEO,MUSCLE-VCD,TRECVID} or text queries \cite{msvd, lsmdc, activitynet, msrvtt, didemo, vatex}. Video queries work better for details hard to describe in words, enabling precise searches. 
% Text queries help focus on key content by ignoring irrelevant visuals. Therefore, we propose the composed query by using video as the dominant content, supplemented by text description to highlight the key content. 
% In order to refine the search scope, we further introduce tag prompts to filter videos by required or excluded tags. Besides, we add mask prompts to improve the referential ability of text description and tag prompts. 
% Recently, \textcolor{red}{composed video retrieval (CVR) \cite{covr, thawakar2024composedvideoretrievalenriched} uses a reference image/video and a text modification as composed query.} It only handles trimmed videos and simple one-to-one video modification. In comparison, our setting is designed for untrimmed videos with challenging tags and one-to-many video retrieval and filtering. 

\textbf{Comprehensive Evaluation Criteria.}
Following the query and partition designs above, we create three versions of the MUVR benchmark. The basic version, MUVR-Base, contains 53K user-uploaded videos from the video platform Bilibili. It includes five partitions, 1,050 video queries with text descriptions, and 84K labeled positive matches. MUVR-Filter further annotates the positive samples with 74K multi-labeled tags. Based on these tags, tag prompts are constructed to filter new positives. We evaluate 3 state-of-the-art video retrieval models \cite{wang2024internvideo2scalingfoundationmodels,s2vs,covr} and 6 Vision-Language Models (VLMs) \cite{evaclip, clip,openclip,blip,blip2} on MUVR-Base and MUVR-Filter. The powerful EVA-CLIP \cite{evaclip} only achieves 58\% and 34\% mAP, respectively, showing limitations in processing untrimmed videos and multi-modal queries. We further build MUVR-QA with 200 query-target relevance judgment questions based on hard samples and design a Reranking Score to evaluate the reranking ability. We evaluate 10 Multi-modal Large Language Models (MLLMs) \cite{internvl25,minicpm,minicmpv,llavanext,llavaonevision,llavavideo} on MUVR-QA. While MLLMs achieve a discrimination accuracy of above 60\%, the Reranking Score indicates that current MLLMs are not yet reliable for reranking tasks. 
% Detailed benchmark information is provided in Section~\ref{sec:MUVR Benchmark}.
%During experiments, we find that dense frame sampling introduces noise, so we propose a frame selection scoring strategy for optimization. 
In summary, our contributions are as follows:

1) We propose Multi-modal Untrimmed Video Retrieval, along with a benchmark MUVR. It features a practical retrieval paradigm with a video-centric multi-modal query format. Five partitions are constructed to cover diverse video categories on the Internet based on multi-level visual correspondence.

2) We evaluate 3 state-of-the-art video retrieval models and 6 VLMs on MUVR-Base/Filter, conducting a thorough analysis of their capabilities across different partitions and query formats. We further propose a Reranking Score to assess the reranking capability of 10 MLLMs by MUVR-QA.

3) We reveal specific limitations in video retrieval methods and VLMs for handling untrimmed videos and multimodal queries, as well as limitations in MLLMs for multi-video understanding and reranking. Inspire future video retrieval research.

%3) We propose a frame selection scoring strategy to reduce noise introduced by dense frame sampling, achieving consistent improvements for both VLMs and VLLMs.
\begin{table}[!t]
\centering
\caption{Definition of visual correspondence between $S_q$ and $S_t$. $S_q$/$S_t$: Any segment of the query/target video.}
\begin{tabular}{|l|p{0.8\textwidth}|}
\hline
\textbf{Copy}       & $S_t$ is copied/edited from $S_q$.\\ \hline
\textbf{Scene}      & $S_q$ and $S_t$ share the same scene/background/region.\\ \hline
\textbf{Instance}   & $S_q$ and $S_t$ share the same instance/object.\\ \hline
\textbf{Action}     & $S_q$ and $S_t$ share the same human action.\\ \hline
\textbf{Event}      & $S_q$ and $S_t$ share the same event with spatio-temporal intersection. \\ \hline
\textbf{Others}     & $S_q$ and $S_t$ are relevant for any of the above correspondence or subjective feeling.\\ \hline
\end{tabular}
\label{tab: Definition of visual correspondence}
\end{table}

\begin{figure*}[t]
    \centering
    \includegraphics[width=1\linewidth]{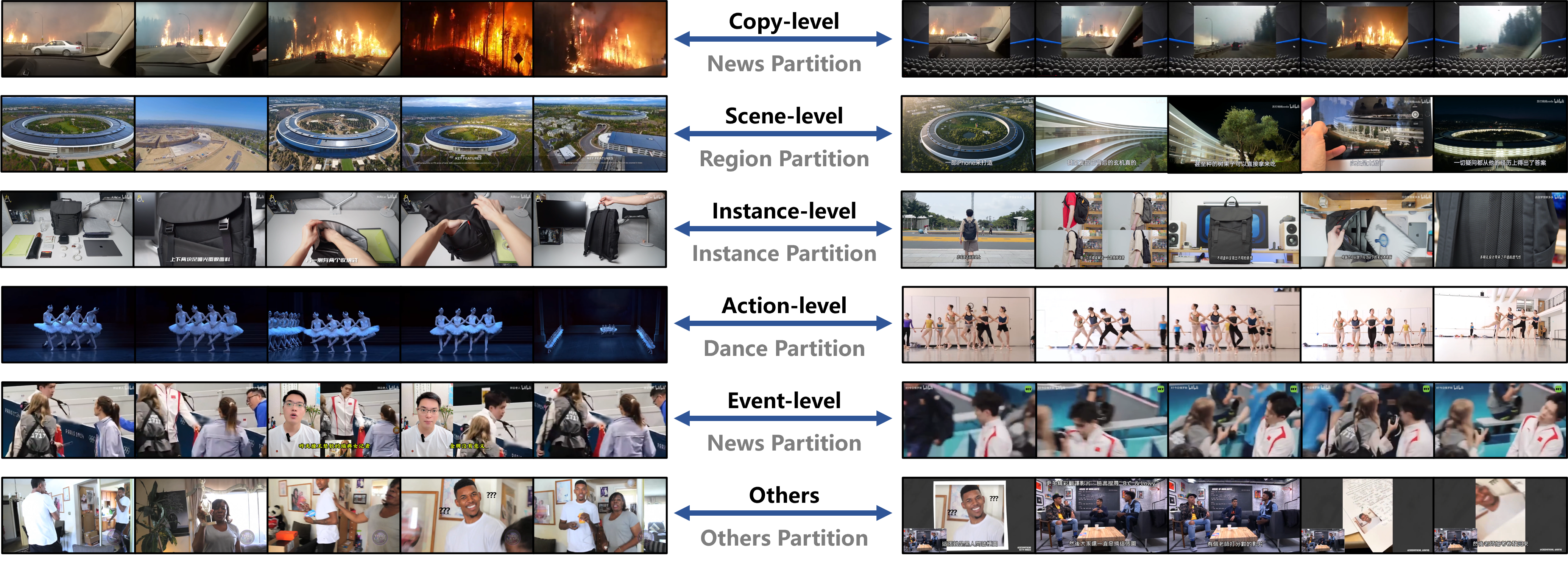}
    \caption{Visualization of videos from different partitions of MUVR. The two videos are matched due to multi-level visual correspondence. Please refer to the Appendix for more results.}
    \label{fig: visualization}
\end{figure*}

\section{Related Work}
\textbf{Video Retrieval.} Text-Video Retrieval (TVR) \cite{msvd, lsmdc, activitynet, msrvtt, didemo, vatex} aims to retrieve relevant videos based on text queries, while Composed Video Retrieval (CVR) \cite{covr,thawakar2024composedvideoretrievalenriched} focuses on retrieving modified videos using a reference image/video along with a text modification of the desired changes. Since most existing benchmarks for these tasks \cite{activitynet,didemo,covr} are derived from video captioning datasets or LLM annotations, they primarily focus on trimmed videos and one-to-one retrieval. However, real-world web video retrieval often involves untrimmed videos and one-to-many retrieval, where relying solely on text queries proves insufficient.
Unlike TVR and CVR, Fine-grained Video Retrieval (FVR) \cite{evve, fivr} focuses on searching untrimmed videos using a video query. Related benchmarks include \cite{evve, CC_WEB_VIDEO, fivr, MUSCLE-VCD, TRECVID, UQ_VIDEO, vcdb}, which typically consist of web video clips uploaded by users on video platforms like YouTube, Google Video, and Yahoo Video, as well as TV shows and movies. Among them, \cite{CC_WEB_VIDEO, UQ_VIDEO} specializes in near-duplicate video retrieval, while \cite{MUSCLE-VCD, TRECVID, vcdb} focuses on video copy detection, both operating at the frame level. Particularly, \cite{evve, fivr} introduced event-level retrieval, requiring target videos to contain events that are spatially and temporally close to those in the query video. Although built upon complex web videos, these benchmarks only support event-level retrieval for the news and film categories. To address this limitation, we propose multi-level visual correspondences for multi-level retrieval to significantly enhance applicability. Furthermore, our benchmark features more diverse video content and categories with more recent publication years, better aligning with real-world web video retrieval scenarios.

\textbf{Multimodal Large Language Models for Reranking.} With the advancement of multimodal retrieval and Multimodal Large Language Models (MLLMs)\cite{yang2025omgmorchestratemultiplegranularities,chen2024mllmstrongrerankeradvancing,khan2025vrragopenvocabularyspeciesrecognition,nguyennhu2025lightweightmomentretrievalglobal,li2025semcoresemanticenhancedgenerativecrossmodal,ling2025mmkbragmultimodalknowledgebasedretrievalaugmented,yeo2025universalragretrievalaugmentedgenerationmultiple,fragomeni2025leveragingmodalitytagsenhanced,wen2025rsragbridgingremotesensing, tsoi2025crossmusimcrossmodalframeworkmusic}, some studies have adopted a two-stage retrieval workflow \cite{vendrow2024inquirenaturalworldtexttoimage,wei2023uniirtrainingbenchmarkinguniversal}. In the first stage, a similarity score is computed based on pre-extracted embeddings to efficiently obtain initial rankings. The second stage applies more costly but sophisticated MLLMs to rerank the top retrievals. For instance, INQUIRE \cite{vendrow2024inquirenaturalworldtexttoimage} prompts MLLMs with ``Does this image show {query}? Answer with `Yes' or `No' and nothing else.'' for text-to-image retrieval, while M-BEIR \cite{wei2023uniirtrainingbenchmarkinguniversal} prompts MLLMs with ``{query}{target} Does the above two images have the same scene? True or False'' for image-to-image retrieval. However, due to the complexity of videos, this workflow has not been thoroughly explored in FVR applications. Consequently, WebVR-QA introduces a query-target relevance discrimination task along with a Reranking Score to evaluate the reranking capability of MLLMs for FVR tasks.

\begin{figure*}[!t]
    \centering
    \includegraphics[width=1\linewidth]{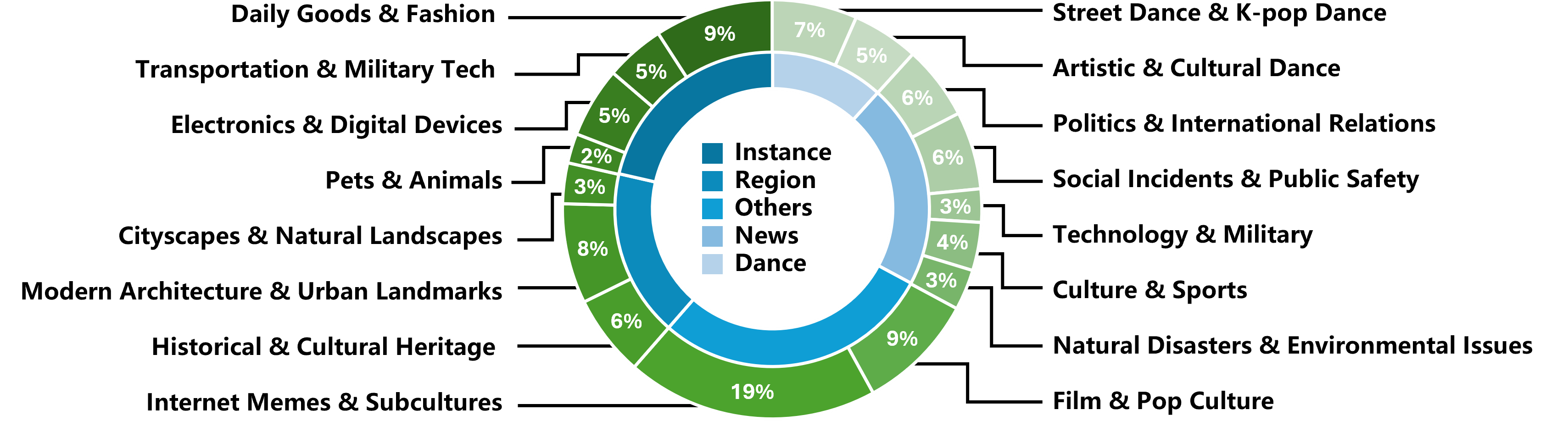}
    \caption{Illustration of video category breakdown for five partitions that make up MUVR.}
    \label{fig: category}
\end{figure*}

\begin{table}[!t]
\centering
\caption{Characteristics of five partitions of MUVR. Each partitions differ in four aspects (i.e., video category, key component of videos, user retrieval interests, and visual correspondence).}
\small
% \footnotesize
\setlength{\tabcolsep}{2mm}
\begin{tabular}{lllcl}
\toprule
Partition & Category & Key Component & User Retrieval Interests & Correspondence \\
\midrule
News & news & frame, event & specific news video clips/frames & copy, event \\
Region & travel vlog etc. & scene & location of video shooting & scene \\
Instance & goods, pets etc. & instance & special objects in the video & instance \\
Dance & dance & action & the actions of people in the video & action \\
Others & meme, film etc. & comprehensive & popular elements in videos & others \\
\bottomrule
\end{tabular}
\label{tab: five partitions of MUVR}
\end{table}

\section{MUVR Benchmark}
\label{sec:MUVR Benchmark}

Here we describe our multi-partition benchmark MUVR for Multi-modal Untrimmed Video Retrieval. MUVR contains 1,050 video-centric multi-modal queries, each comprising a video query with text descriptions, tag prompts, and mask prompts. These queries are mapped to relevant matches across 53,462 videos collected from the video platform Bilibili, covering diverse video categories and organized into five partitions based on our proposed multi-level visual correspondence. 
%Figure \ref{fig1} shows example queries, with additional cases in the supplementary material. 
This section explains the benchmark construction process, annotation methodology, and evaluation protocols.

\subsection{Composition and Collection}
\label{sec:Composition and Collection}

\textbf{Definition of Visual Correspondence and Partitions.} Key components of a video mainly include low-level frames, mid-level semantics such as scene, instance and action, as well as high-level semantics such as event and others. These are also what users tend to be interested in and want to retrieve, corresponding to six levels of visual correspondence (i.e., copy, scene, instance, action, event, and others) as shown in Table \ref{tab: Definition of visual correspondence}. Based on the multi-level visual correspondence, MUVR organizes videos into five partitions (i.e., news, region, instance, dance, and others) as shown in Table \ref{tab: five partitions of MUVR}. Figure \ref{fig: category} illustrates the video category breakdown for five partitions. MUVR covers diverse common video categories in Bilibili and is compliant with partition characteristics for comprehensive evaluation. 
%To ensure content diversity, MUVR contains most of the video taxonomy of Bilibili and includes challenging formats like vlogs, product demos, dance videos, and fan-made content.

\textbf{Topics and Video Collection.} To efficiently collect various videos of different categories and contents, we design 350 search topics based on trending keywords of Bilibili and split them into five partitions. The search topics should meet three criteria: 1) sufficient relevant videos to retrieve on the platform; 2) distinctive visual content in relevant videos, and 3) visual uniqueness compared to videos of other topics. Subsequently, we collect the top 100 search results per topic, remove videos exceeding 6 minutes, and then crop long videos to 2-minute videos (untrimmed) following \cite{evve,fivr}. Final processing includes resizing to 336 pixels (long edge) and downsampling to 6 fps.

\textbf{Query Collection and Annotation.} The collection of queries and the annotation process are completed by professional annotators. They are first instructed to analyze the key visual content of videos of each topic. For each topic, three representative and visually distinct videos are chosen as video queries. For each query, detailed text descriptions are created to specify key visual contents and retrieval needs. Annotations are restricted to videos within the same topic, as cross-topic videos are deemed irrelevant based on topic differences and video similarity screening (we calculate the similarity score with BLIP2-features and guarantee that the query video is unrelated to the 10 most similar videos from other topics). To ensure consistency, all videos undergo two rounds of annotation, and those with conflicting labels across rounds are excluded. This process produces MUVR-Base, where each query has about 80 verified positive matches.

\begin{wraptable}{r}{0.7\textwidth} % Adjust position (r/l) and width as needed
\centering
\caption{Statistics of MUVR-Filter (based on MUVR-Base) with different Tag Prompt formats.}
\small
\setlength{\tabcolsep}{1mm}
\label{tab: tag comparison}

\begin{tabular}{lcrrr}
    \toprule
    Dataset & Tag Prompt & Queries & Matches & Positive Rate \\
    \midrule
    MUVR-Base & - & 1,050 & 84,035 & 55.2\% \\
    MUVR-Filter & ``$\pm$ [tag]'' & 9,979 & 385,818 & 27.6\% \\
    \midrule
    \begin{tabular}[l]{@{}l@{}}MUVR-Filter\\ (upper bound)\end{tabular} & \begin{tabular}[c]{@{}c@{}c@{}}``$\pm$ [tag]'' \\ ``$\pm$ [tag] AND $\pm$ [tag]'' \\ ``$\pm$ [tag] OR $\pm$ [tag]''\end{tabular} & 93,885 & 4,284,265 & 29.4\% \\
    \bottomrule
\end{tabular}
\end{wraptable}

\textbf{Tag Prompt and Mask Prompt.} Based on MUVR-Base, annotators are instructed to design 3\textasciitilde10 tags per topic. These tags capture shared attributes among positive retrieval samples, such as challenging video styles (e.g., animation vs. live-action), camera perspectives (e.g., first-person view), and domain variations (e.g., outdoor vs. indoor). Tags are optionally assigned to both query videos and their matches (zero or multiple tags per item). These tags enabled hierarchical filtering through tag prompts as shown in Table \ref{tab: tag comparison}, forming our MUVR-Filter with 9,979 queries. Additionally, we enhance the text descriptions' fine-grained reference representation to video queries by adding mask prompts with SAM2 \cite{ravi2024sam2segmentimages}.

\textbf{MUVR-QA.} We construct 200 challenging discrimination questions to evaluate MLLMs with query-target cases where EVA-CLIP \cite{evaclip} fails on MUVR-Base and MUVR-Filter. Specifically, for each query with mAP below 0.05, we select two targets: the highest-scoring true match and the highest-scoring false match. This generates 102 questions from MUVR-Base and 98 from MUVR-Filter.

\begin{table*}[!t]
\caption{Comparison of MUVR-Base with the most related FVR benchmarks. ${\ddagger}$: construct using video transformations. ${\dagger}$: expand with hundreds of thousands of unrelated videos. YT: YouTube. TV: TV show. B: Bilibili. C: Copy. E: Event. S: Scene. I: Instance. A: Action. O: Others.}
    \centering
    \setlength{\tabcolsep}{1mm}
    \begin{tabular}{lrrrrrrcc}
        \toprule
        Benchmarks & Topics & Queries & Matches & Videos & Hours & \begin{tabular}[c]{@{}c@{}}Source\\(Year)\end{tabular} & Category & \begin{tabular}[c]{@{}c@{}}Corres-\\ pondence\end{tabular}  \\
        
        \midrule
        
        CC\_WEB\_VID\cite{CC_WEB_VIDEO} & 24 & 24 & 3,481 & 12,790 & 551 & YT(06) & mixed & C  \\
        UQ\_VIDEO\cite{UQ_VIDEO} & 24 & 24 & 3,481 & 169,952$^{\dagger}$ & N/A & YT(09) & mixed & C  \\
        MUSCLE-VCD\cite{MUSCLE-VCD} & 15 & 15 & N/A & 101 & 100 & TV(07) & film & C  \\
        TRECVID\cite{TRECVID} & N/A & 11,256$^{\ddagger}$ & N/A & 11,503 & 420 & TV(11) & film  & C  \\

        VCDB\cite{vcdb} & 28 & 528 & 9,236 & 100,528$^{\dagger}$ & 2,038 & YT(14) & \begin{tabular}[c]{@{}c@{}}news, film\end{tabular} & C  \\

        EVVE\cite{evve} & 13 & 620 & 1,252 & 102,375$^{\dagger}$ & 5,536 & YT(12) & news & E  \\
        
        FIVR\cite{fivr} & 100 & 100 & 12,300 & 225,960$^{\dagger}$ & 7,100 & YT(17) & news & C, E  \\
        
        \midrule
        MUVR-Base (Ours) & 350 & 1,050 & 84,035 & 53,462 & 1,762 & B(24) & \begin{tabular}[c]{@{}c@{}}multi- \\partition\end{tabular} & \begin{tabular}[c]{@{}c@{}}C, E, S \\I, A, O \end{tabular}  \\
        \bottomrule
    \end{tabular}
    
    \label{tab: FVR dataset comparison}
\end{table*}

% \begin{table*}[!t]
% \caption{Partition statistics of MUVR-Base\&Tag. Ent.: Entertainment. Des.: words of description.}
%     \centering
%     \setlength{\tabcolsep}{1mm}
%     \begin{tabular}{lrrrrrrrccc}
%         \toprule
%         \multirow{2}{*}{Partition} & \multirow{2}{*}{Topics} & \multicolumn{2}{c}{\textbf{MUVR-Base}} & \multicolumn{2}{c}{\textbf{MUVR-Filter}} & \multirow{2}{*}{Videos} & \multirow{2}{*}{Tags} & Tag & Des. & Mask \\
%         \cmidrule(lr){3-4} \cmidrule(lr){5-6}
%         & & Queries & Matches & Queries & Matches & & & Labels & (Avg.) & Prompts\\
%         \midrule
%         News & 74 & 222 & 12,273 & 2,304 & 70,876 & 9,993 & 474 & 14,964 & 20 & 169  \\
%         Region & 60 & 180 & 14,544 & 1,960 & 79,596 & 9,005 & 349 & 11,158 & 14 & 111  \\
%         Instance & 75 & 225 & 23,844 & 2,213 & 119,496 & 13,009 & 390 & 15,738 & 15 & 132  \\
%         Dance & 41 & 123 & 11,687 & 1,222 & 15,525 & 7,026 & 165 & 10,054 & 27 & 97  \\
%         Ent. & 100 & 300 & 21,687 & 2,280 & 100,325& 14,429 & 475 & 22,365 & 26 & 283  \\
%         \midrule
%         Overall & 350 & 1,050 & 84,035 & 9,979 & 385,818 & 53,462 & 1,853 & 74,279 & 20 & 792 \\
%         \bottomrule
%     \end{tabular}
    
%     \label{tab: partition statistics}
% \end{table*}

\begin{table*}[!t]
\caption{Partition statistics of MUVR-Base\&Tag. Des.: words of description.}
    \centering
    \small
    \setlength{\tabcolsep}{1mm}
    \begin{tabular}{lrrrrrrrccc}
        \toprule
        \multirow{2}{*}{Partition} & \multirow{2}{*}{Topics} & \multicolumn{2}{c}{\textbf{MUVR-Base}} & \multicolumn{2}{c}{\textbf{MUVR-Filter}} & \multirow{2}{*}{Videos} & \multirow{2}{*}{Tags} & Tag & Des. & Mask \\
        \cmidrule(lr){3-4} \cmidrule(lr){5-6}
        & & Queries & Matches & Queries & Matches & & & Labels & (Avg.) & Prompts\\
        \midrule
        News & 74 & 222 & 12,273 & 2,304 & 70,876 & 9,993 & 474 & 14,964 & 20 & 20  \\
        Region & 60 & 180 & 14,544 & 1,960 & 79,596 & 9,005 & 349 & 11,158 & 14 & 20  \\
        Instance & 75 & 225 & 23,844 & 2,213 & 119,496 & 13,009 & 390 & 15,738 & 15 & 20  \\
        Dance & 41 & 123 & 11,687 & 1,222 & 15,525 & 7,026 & 165 & 10,054 & 27 & 20  \\
        Others & 100 & 300 & 21,687 & 2,280 & 100,325& 14,429 & 475 & 22,365 & 26 & 20  \\
        \midrule
        Overall & 350 & 1,050 & 84,035 & 9,979 & 385,818 & 53,462 & 1,853 & 74,279 & 20 & 100 \\
        \bottomrule
    \end{tabular}
    
    \label{tab: partition statistics}
\end{table*}

% \begin{table*}[!t]
% \caption{Statistics of MUVR-Filter (based on MUVR-Base) with different Tag Prompt formats.}
%     \small
%     \setlength{\tabcolsep}{1mm}
%     \centering
%     \begin{tabular}{lcrrr}
%         \toprule
%         Dataset & Tag Prompt & Queries & Matches & Positive Rate \\
%         \midrule
%         MUVR-Base & - & 1,050 & 84,035 & 55.2\% \\
%         MUVR-Filter & ``$\pm$ [tag]'' & 9,979 & 385,818 & 27.6\% \\
%         \midrule
%         \begin{tabular}[l]{@{}l@{}}MUVR-Filter\\ (upper bound)\end{tabular} & \begin{tabular}[c]{@{}c@{}c@{}}``$\pm$ [tag]'' \\ ``$\pm$ [tag] AND $\pm$ [tag]'' \\ ``$\pm$ [tag] OR $\pm$ [tag]''\end{tabular} & 93,885 & 4,284,265 & 29.4\% \\

%         \bottomrule
%     \end{tabular}
    
%     \label{tab: tag comparison}
% \end{table*}

\textbf{Statistics.}
As shown in Table \ref{tab: FVR dataset comparison}, MUVR-Base features richer topics, queries, video categories, and visual correspondences. Note that the video queries in \cite{TRECVID} are artificially generated through editing modifications rather than from original videos. \cite{UQ_VIDEO, vcdb, evve, fivr} contains hundreds of thousands of simple, irrelevant videos, increasing evaluation overhead. Furthermore, previous benchmarks consist of web videos from before 2017 with limited video content and categories. 
Table \ref{tab: partition statistics} displays statistics across different partitions of MUVR, where our text descriptions average 20 words for fine-grained representation. As shown in Table \ref{tab: tag comparison}, tag prompt significantly reduces the positive rate of queries for videos from the same topic. When applying binary relations to construct the Tag Prompt, 93,885 queries and 4M matches can be obtained, indicating the flexibility of finer retrieval matching based on tags.

% \textbf{Video analysis.}
% 视频数量、时长、画面尺寸、fps
% 正样本比例

% \textbf{Text query analysis.}
% text query、tag prompt数量、平均长度
% 文本语义类型、词云（附录）

\subsection{Evaluation Metric}

Following \cite{fivr, covr}, MUVR-Base is evaluated using mAP, uAP, and Recall at k (R@k). MUVR-Filter is evaluated using mAP and Recall at k (R@k). Additionally, MUVR-QA is evaluated using Accuracy and our proposed Reranking Score.

\textbf{Reranking Score.} This metric simulates real-world retrieval reranking scenarios where true positives should be preserved and false positives removed. Each MUVR-QA query contains two targets: a true positive (label 1) and a false positive (label 0). When processing these pairs (label 10), a VLLM may produce four kinds of outcome: \textbf{10} (correctly keeps true and removes false), \textbf{11} (retains both, equivalent to no reranking action), \textbf{00} (incorrectly removes both), or \textbf{01} (wrongly removes true while keeping false). We assign scores of +1, 0, -1, and -2, respectively, where outcome \textbf{11} receives a score of 0 because it maintains the original retrieval result unchanged. The final Reranking Score averages these values across all queries, measuring models' ability to refine initial rankings.

\section{Experiments}
\label{sec: experiments}
\subsection{Models and Methods for Evaluation}
\label{sec: Models and Methods for Evaluation}
We describe the models and methods evaluated on MUVR. 
On MUVR-Base and MUVR-Filter, we evaluate 3 state-of-the-art video retrieval models \cite{wang2024internvideo2scalingfoundationmodels,s2vs,covr} and 6 image-based VLMs \cite{evaclip, clip,openclip,blip,blip2}. Specifically, we assess the retrieval performance for `each partition' with the corresponding video library and report the average result. The retrieval queries on MUVR-Base include three formats: pure text description, pure video query, and the combination of both (multimodal query). On MUVR-QA, we evaluate the query-target video relevance discrimination capabilities of 10 open-source MLLMs \cite{internvl25,minicpm,minicmpv,llavanext,llavaonevision,llavavideo,yuan2025videorefersuite,openai2024gpt4o,gemini}. All the experiments are conducted on a workstation with 8 Tesla V100 GPUs.

\textbf{Video Retrieval.} We first introduce 3 video retrieval models. S2VS \cite{s2vs} is trained on 100K videos \cite{dns} and achieves state-of-the-art performance on three video-to-video retrieval datasets \cite{vcdb,evve,fivr}. The large-scale pre-trained model InternVideo2 \cite{wang2024internvideo2scalingfoundationmodels} obtains state-of-the-art results on multiple text-to-video retrieval datasets \cite{msvd, lsmdc, activitynet, msrvtt, didemo, vatex}. For composed video retrieval methods, we select CoVR \cite{covr} because it doesn't rely on additional information of target videos, and its training parameter is open-sourced. Given the challenging nature of MUVR, we further introduce 6 VLMs pre-trained on massive image-text datasets for evaluation. Specifically, we uniformly sample $N=15$ frames from the video $V$ to extract feature $VLM(V)\in R^{N*d}$ with dimension $d$. For text $T$, we extract feature $VLM(T)\in R^{d}$. We calculate the similarity matrix and extract the maximum value as the retrieval score as follows:
\begin{equation*}
\text{Score}(a, b) = \max( VLM(a) VLM(b)^\top ),
\end{equation*}
\begin{equation*}
S_v = \text{Score}(V_{\text{query}}, V_{\text{target}}),\ \ S_t = \text{Score}(T_{\text{description}}, V_{\text{target}}),
\end{equation*}
\begin{equation*}
S_{tv} = (S_t + S_v)/2,\ \ S_{tag} = S_{tv} + p \times \text{Score}(T_{\text{tag}}, V_{\text{target}}),
\end{equation*}

% \begin{equation*}
% \text{Score}\left(a, b\right) = \max\left( VLM(a) VLM(b)^\top \right), S_v=\text{Score}\left(V_{\text{query}}, V_{\text{target}}\right), 
% \end{equation*}
% \begin{equation*}
% S_t=\text{Score}\left(T_{\text{description}}, V_{\text{target}}\right),  S_{tv}=\left(S_t + S_v\right)/2, S_{tag}=S_{tv} + p \times \text{Score}\left(T_{\text{tag}}, V_{\text{target}}\right),
% \end{equation*}

where $p=\pm0.3$ according to the sign of the Tag Prompt.

% \textbf{Frame Selection Scoring Strategy.}
% 介绍如何减少无关帧的干扰，画一张图对比优化前后的方法

% TBD

\begin{figure*}[!t]
    \centering
    \includegraphics[width=1\linewidth]{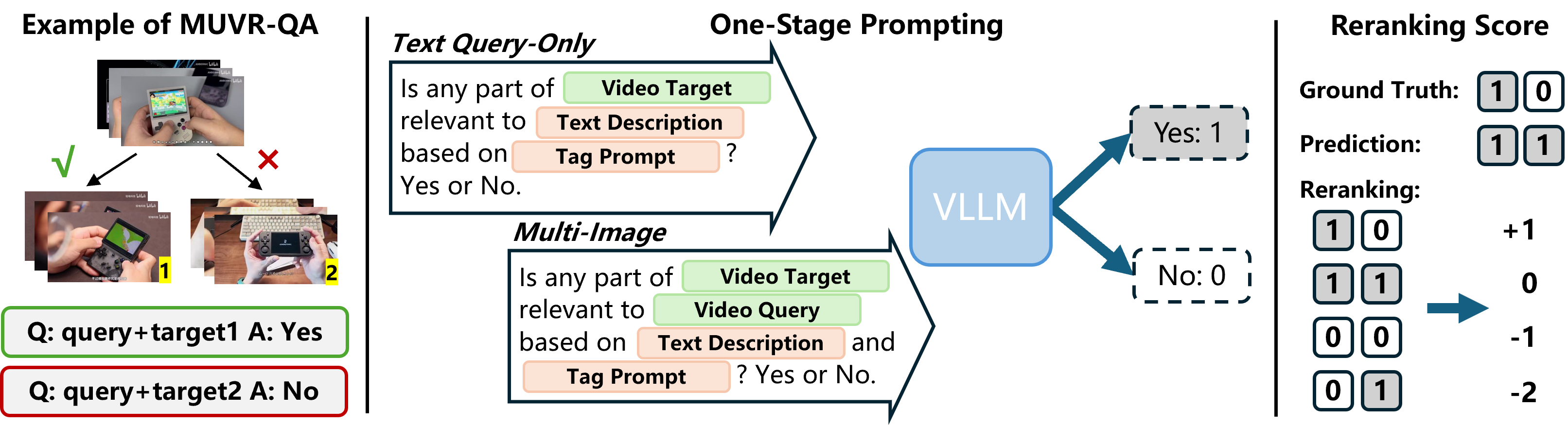}
    \caption{Illustration of MUVR-QA, Reranking Score, and MLLMs prompting.}
    \label{fig4}
\end{figure*}

\textbf{MLLMs for MUVR-QA.} Some image-based multi-modal retrieval methods \cite{vendrow2024inquirenaturalworldtexttoimage,wei2023uniirtrainingbenchmarkinguniversal} have explored using MLLMs to assess relevance between the image-text query and the target image, showing potential for reranking improvement. 
%However, existing MLLMs lack specialized optimization for processing multiple videos, making it difficult to directly input two videos. Additionally, strategies relying on multi-step dialogues or caption generation are inefficient in practice. 
Consequently, we propose two one-stage approaches that prompt MLLMs to output Yes/No responses to minimize latency. As illustrated in Figure \ref{fig4}, one method feeds text description, tag prompt, and the target video for comparison. Another method employs MLLMs with multi-image understanding capabilities by jointly inputting both the query and target video frames, with explicit prompts indicating which frames originate from which video.

\subsection{Results and Analysis}

\textbf{Retrieval with video and text queries.}
We report the evaluation results of 3 state-of-the-art video retrieval methods and 6 VLMs on MUVR-Base in Table \ref{tab: MUVR-Base combined results} and have the following findings: 

\textbf{Finding 1: The average performance mainly depends on the number of parameters, training data, and model structure.} The performance of CLIP-based models \cite{clip, openclip} improves with stronger backbone (from ResNet50 to ViT-H-14). EVA-CLIP achieves the best results on nearly all metrics, benefiting from its larger parameter count and training data volume. 
When using pure video as the query, the video retrieval models InternVideo2 and S2VS achieve suboptimal mAP and the best uAP, indicating that video-based model structures excel at capturing inter-frame temporal relationships of videos. 
When using pure text description as the query, image-based VLMs \cite{openclip, evaclip} perform the best, mainly because their training data contains more image-text pairs uploaded by users in online communities, which aligns with video platform content. 

\textbf{Finding 2: A reasonable video retrieval framework combined with a powerful image backbone can achieve improvement.}
The fine-grained incident video retrieval model S2VS is trained on CLIP (RN50x4) features and achieve great improvement (from 34.2\% to 47.2\% on mAP and from 16.1\% to 36.6\% on uAP), demonstrating that pre-extracted frame features from VLMs can be further improved through video retrieval framework training. In contrast, CoVR only supports 1-frame query video input, making it less suitable for untrimmed video and thus resulting in poorer performance. 

\textbf{Finding 3: Video query is more important, and multimodal query further improves the performance.}
Using pure video as a query generally yields better performance than pure text, as video queries can more precisely represent details that are difficult to describe in text. However, for the Instance partition, the key component is usually small, and pure video queries introduce more irrelevant content interference, leading to poorer model performance compared to pure text queries. Using multimodal queries can often achieve significant improvements, indicating the complementary role of video and text in retrieval. However, in some cases (e.g., the mAP of InternVdeo2 on News partition), the improvement of multimodal queries is limited, which inspires us to explore more effective methods for understanding multimodal queries. 

\textbf{Finding 4: Videos from different partitions have different key content and visual correspondence, posing challenges to the generality of current methods.}
EVA-CLIP performs best in the News, Instance, and Region partitions. This is because retrieval in the Instance and Region partitions emphasizes static spatial understanding, which VLMs excel at, and retrieval in the News partition primarily relies on instances and scenes within videos. The video models InternVideo2 and S2VS perform best in the Others and Dance partitions, indicating that retrieval in these two partitions relies more on dynamic temporal understanding, such as coherent movements or continuous narratives. 

\begin{table*}[!t]
\centering
\caption{Retrieval performance on MUVR-Base with video and text queries. N: News. O: Others. I: Instance. R: Region. D: Dance.}
\label{tab: MUVR-Base combined results}
\small
\setlength{\tabcolsep}{1mm}
\begin{tabular}{lcccccccccc}
\toprule
 \multirow{2}{*}{\textbf{Method}} & \multicolumn{5}{c}{\textbf{average performance}} & \multicolumn{5}{c}{\textbf{mAP of different partitions}} \\
 \cmidrule(lr){2-6} \cmidrule(lr){7-11}
 & \textbf{mAP} & \textbf{uAP} & \textbf{R@200} & \textbf{R@500} & \textbf{R@2000} & \textbf{N} & \textbf{O} & \textbf{I} & \textbf{R} & \textbf{D} \\
\hline
\rowcolor{blue!10}
\multicolumn{11}{c}{\textbf{\textit{Pure Text as Query}}} \\

CLIP (RN50x4)\cite{clip} & 27.7 & 14.7 & 45.8 & 60.6 & 81.0 & 28.7 & 26.4 & 41.0 & 33.2 & 9.3 \\
CLIP (ViT-L/14@336px)\cite{clip} & 35.0 & 21.4 & 54.0 & 67.2 & 83.7 & 38.5 & 33.9 & 47.8 & 41.5 & 13.6 \\
OpenCLIP (ViT-H-14)\cite{openclip} & 39.7 & 20.3 & 58.3 & 70.6 & 85.6 & 41.1 & 40.3 & 55.8 & 44.9 & 16.1 \\
EVA-CLIP\cite{evaclip} & 43.0 & 23.3 & 61.9 & 73.3 & 86.9 & 45.1 & 44.1 & 59.7 & 49.9 & 16.3 \\
BLIP\cite{blip} & 28.0 & 12.9 & 46.7 & 60.7 & 80.2 & 27.5 & 26.6 & 41.4 & 33.7 & 10.9 \\
BLIP2\cite{blip2} & 31.3 & 18.0 & 50.4 & 64.4 & 82.6 & 32.0 & 31.8 & 45.9 & 35.9 & 10.8 \\
InternVideo2\cite{wang2024internvideo2scalingfoundationmodels} & 36.9 & 23.2 & 55.0 & 68.5 & 85.2 & 38.0 & 44.8 & 44.8 & 40.8 & 16.4 \\

\hline
\rowcolor{blue!10}
\multicolumn{11}{c}{\textbf{\textit{Pure Video as Query}}} \\

CLIP (RN50x4)\cite{clip} & 34.2 & 16.1 & 48.6 & 61.7 & 82.1 & 40.7 & 46.4 & 31.8 & 32.4 & 19.5 \\
CLIP (ViT-L/14@336px)\cite{clip} & 38.4 & 18.9 & 52.8 & 64.7 & 82.8 & 48.6 & 48.9 & 36.4 & 35.6 & 22.4 \\
OpenCLIP (ViT-H-14)\cite{openclip} & 46.0 & 29.9 & 61.7 & 73.2 & 87.7 & 52.6 & 55.8 & 47.0 & 49.6 & 24.9 \\
EVA-CLIP\cite{evaclip} & 50.7 & 33.1 & 66.6 & 77.8 & 90.5 & 57.8 & 59.2 & 54.7 & 55.1 & 26.9 \\
BLIP\cite{blip} & 35.5 & 17.4 & 49.3 & 61.4 & 80.2 & 45.1 & 46.7 & 33.1 & 34.2 & 18.3 \\
BLIP2\cite{blip2} & 46.0 & 30.0 & 61.1 & 73.5 & 88.5 & 54.1 & 55.8 & 47.7 & 46.6 & 25.6 \\
InternVideo2\cite{wang2024internvideo2scalingfoundationmodels} & 48.0 & 36.9 & 62.9 & 75.2 & 89.4 & 56.1 & 62.0 & 47.5 & 45.7 & 28.5 \\
S\textsuperscript{2}VS\cite{s2vs} & 47.2 & 36.6 & 60.4 & 70.2 & 84.6 & 51.3 & 63.7 & 49.5 & 49.1 & 22.5 \\

\hline
\rowcolor{blue!10}
\multicolumn{11}{c}{\textbf{\textit{Multimodal Query}}} \\

CLIP (RN50x4)\cite{clip} & 42.9 & 29.8 & 58.8 & 71.1 & 87.0 & 49.4 & 53.6 & 46.5 & 43.8 & 21.2 \\
CLIP (ViT-L/14@336px)\cite{clip} & 49.2 & 35.7 & 64.4 & 75.6 & 88.6 & 58.2 & 57.7 & 54.3 & 50.7 & 25.2 \\
OpenCLIP (ViT-H-14)\cite{openclip} & 54.0 & 40.1 & 69.2 & 79.3 & 90.8 & 59.4 & 62.7 & 62.3 & 59.1 & 26.7 \\
EVA-CLIP\cite{evaclip} & 58.0 & 44.6 & 73.0 & 82.5 & 92.3 & 63.1 & 66.1 & 68.2 & 63.8 & 28.7 \\
BLIP\cite{blip} & 44.1 & 29.5 & 59.4 & 71.2 & 86.2 & 50.3 & 54.2 & 59.4 & 47.0 & 19.4 \\
BLIP2\cite{blip2} & 51.0 & 38.7 & 66.5 & 77.3 & 89.7 & 56.3 & 61.7 & 58.1 & 53.7 & 25.4 \\
InternVideo2\cite{wang2024internvideo2scalingfoundationmodels} & 52.1 & 37.4 & 66.9 & 78.4 & 90.7 & 57.3 & 66.3 & 55.3 & 52.5 & 28.9 \\
CoVR\cite{covr} & 43.3 & 30.9 & 62.1 & 74.9 & 89.2 & 50.5 & 54.3 & 46.9 & 44.0 & 20.8 \\

\bottomrule
\end{tabular}
\end{table*}

\begin{table*}[!t]
\centering
\caption{Retrieval performance on MUVR-Filter with multimodal query and tag prompt. N: News. O: Others. I: Instance. R: Region. D: Dance.}
\label{tab: MUVR-Filter combined results}
\small
\setlength{\tabcolsep}{1mm}
\begin{tabular}{lcccccccccc}
\toprule
 \multirow{2}{*}{\textbf{Method}} & \multicolumn{5}{c}{\textbf{average performance}} & \multicolumn{5}{c}{\textbf{mAP of different partitions}} \\
 \cmidrule(lr){2-6} \cmidrule(lr){7-11}
 & \textbf{mAP} & \textbf{R@200} & \textbf{R@500} & \textbf{R@1000} & \textbf{R@2000} & \textbf{N} & \textbf{O} & \textbf{I} & \textbf{R} & \textbf{D} \\
\hline
\rowcolor{blue!10}
\multicolumn{11}{c}{\textbf{\textit{Multimodal Query without Tag Prompt}}} \\

OpenCLIP (ViT-H-14)\cite{openclip} & 30.8 & 66.5 & 76.9 & 83.6 & 89.5 & 34.6 & 34.3 & 34.9 & 34.1 & 16.2 \\
EVA-CLIP\cite{evaclip} & 32.9 & 70.5 & 80.4 & 86.3 & 91.3 & 36.6 & 36.5 & 37.8 & 36.3 & 17.4 \\
BLIP\cite{blip} & 25.5 & 56.6 & 68.8 & 76.9 & 84.5 & 29.4 & 30.0 & 27.7 & 27.8 & 12.3 \\
BLIP2\cite{blip2} & 29.3 & 63.5 & 74.8 & 81.7 & 88.2 & 33.0 & 34.3 & 32.3 & 31.5 & 15.5 \\

\hline
\rowcolor{blue!10}
\multicolumn{11}{c}{\textbf{\textit{Multimodal Query with Tag Prompt}}} \\

OpenCLIP (ViT-H-14)\cite{openclip} & 31.7 & 66.9 & 77.2 & 83.7 & 89.6 & 36.0 & 35.5 & 34.9 & 35.8 & 16.4 \\
EVA-CLIP\cite{evaclip} & 34.0 & 70.9 & 80.5 & 86.4 & 91.4 & 38.3 & 37.7 & 38.1 & 38.3 & 17.6 \\
BLIP\cite{blip} & 25.6 & 56.4 & 68.3 & 76.5 & 84.4 & 30.4 & 30.0 & 26.6 & 28.7 & 12.1 \\
BLIP2\cite{blip2} & 30.4 & 63.8 & 75.1 & 82.1 & 88.3 & 34.9 & 35.8 & 32.5 & 33.2 & 15.7 \\

\bottomrule
\end{tabular}
\end{table*}

\begin{table*}[!t] % 使用[H]强制在当前位置
\centering
\caption{Performance on MUVR-QA. \textbf{Frame} denotes the number of target video frames. Only the first frame of the query video is sampled as input due to model capability constraints. \textbf{No Tag} represents the subset of MUVR-QA without a tag prompt. \textbf{Delay} is measured per sample on a single V100 GPU or closed source model API. *: using 8-GPU parallel processing. $\dagger$: using mask prompt.}
\small
\label{tab:vllm_MUVR-QA_results}

\definecolor{lightgreen}{RGB}{230,255,230}
\begin{tabular}{lccccccc}
\toprule
\multirow{2}{*}{\textbf{Method}} & \multirow{2}{*}{\textbf{Size}} & \multirow{2}{*}{\textbf{Frame}} & \multicolumn{2}{c}{\textbf{Accuracy}} & \multicolumn{2}{c}{\textbf{Reranking Score}} & \textbf{Delay} \\
\cmidrule(lr){4-5} \cmidrule(lr){6-7}
 & & & \textbf{All} & \textbf{No Tag} & \textbf{All} & \textbf{No Tag} & \textbf{(s)}\\

\hline
\rowcolor{blue!10} % 添加淡蓝色底纹
\multicolumn{8}{c}{\textbf{\textit{One-Stage Text Query-Only Comparison}}} \\

InternVL2\cite{wang2024internvideo2scalingfoundationmodels} & 8B & 6 & 55.0 & 59.8 & -0.52 & -0.31 & 3.56 \\
InternVL2.5\cite{internvl25} & 8B & 1 & 49.0 & 52.9 & -0.84 & -0.65 & 1.12 \\
  &  & 6 & 58.0 & 65.7 & -0.45 & -0.18 & 3.61 \\
  &  & 12 & 58.5 & 61.8 & -0.37 & -0.16 & 7.26 \\

MiniCPM-o 2.6\cite{minicpm} & 8B & 12 & 51.0 & 66.7 & -0.46 & 0.06 & 3.26 \\
MiniCPM-V 2.6\cite{minicmpv} & 8B & 12 & 50.0 & 71.6 & -0.59 & 0.18 & 4.53 \\

LLaVA-NeXT-Video\cite{llavanext} & 7B & 12 & 50.5 & 58.8 & -0.72 & -0.49 & 1.31* \\
LLaVA-OV\cite{llavaonevision} & 7B & 12 & 50.0 & 60.8 & -0.52 & 0.04 & 1.05 \\
LLaVA-Video\cite{llavavideo} & 7B & 12 & 47.0 & 60.8 & -0.38 & 0.08 & 5.38* \\

\hline
\rowcolor{blue!10} % 添加淡蓝色底纹
\multicolumn{8}{c}{\textbf{\textit{One-Stage Multi Image Comparison}}} \\

InternVL2\cite{wang2024internvideo2scalingfoundationmodels} & 8B & 6 & 58.5 & 73.5 & -0.23 & 0.34 & 4.21 \\
InternVL2.5\cite{internvl25} & 8B & 1 & 52.0 & 58.8 & -0.66 & -0.37 & 1.49 \\
  &  & 6 & 57.0 & 71.6 & -0.35 & 0.18 & 4.23 \\
  &  & 12 & 56.5 & 69.6 & -0.33 & 0.15 & 7.73 \\
  
MiniCPM-o 2.6\cite{minicpm} & 8B & 12 & 53.0 & 55.9 & -0.15 & 0.02 & 3.60 \\
MiniCPM-V 2.6\cite{minicmpv} & 8B & 12 & 54.0 & 60.8 & -0.10 & 0.14 & 4.63 \\
VideoRefer\cite{yuan2025videorefersuite} & 7B & 12 & 53.0 & 55.9 & 0.05 & 0.10 & 4.47 \\
VideoRefer$\dagger$\cite{yuan2025videorefersuite} & 7B & 12 & 55.0 & 56.9 & 0.07 & 0.12 & 4.58 \\
Gemini-2.0-Flash\cite{gemini} & N/A & 6 & 60.5 & 62.7 & -0.11 & -0.21 & 3.42 \\
 &  & 12 & 63.5 & 68.6 & 0.07 & 0.16 & 3.77 \\
GPT-4o\cite{openai2024gpt4o} & N/A & 6 & 65.0 & 75.5 & 0.19 & 0.37 & 6.93 \\
 &  & 12 & 62.0 & 69.6 & 0.15 & 0.25 & 8.64 \\

\bottomrule
\end{tabular}
\end{table*}

\textbf{Retrieval with additional tag prompts.}
We report the evaluation results of selected models for MUVR-Filter in Table \ref{tab: MUVR-Filter combined results}. Ignoring the Tag Prompt slightly hurts Recall but significantly reduces mAP, demonstrating the fine-grained retrieval challenge of MUVR-Filter. Although model performance improves with Tag Prompt assistance, there remains significant room for improvement. BLIP shows marginal gains with the combination of Tag Prompts, indicating that small models struggle to comprehend Tag Prompts. Besides, all the models achieve limited improvements in the Instance and Dance partitions, suggesting difficulties in understanding instance-specific terms and dance movements from tags.

\textbf{Reranking Performance of MLLMs.}
% Table \ref{tab:vllm_MUVR-QA_results} presents the performance of 10 MLLMs on MUVR-QA. Overall, questions incorporating Tag Prompts from MUVR-Filter pose challenges for MLLMs. Processing multiple frames significantly improves performance compared to single frames, but increases inference latency (1s vs. 4s). Further increasing the number of frames yields limited Ranking Score improvement and reduced Accuracy, indicating MLLMs' insufficient capability in retrieving key content from dense video frames. Compared to using text queries alone, MLLMs with multi-image understanding capabilities (e.g., InternVL2) can leverage query video information to enhance performance, particularly in the Reranking Score metric. The recent videoRefer can understand mask prompts and has shown better performance with its help, demonstrating the effectiveness of mask prompts.
We evaluate 10 MLLMs on MUVR-QA as shown in Table \ref{tab:vllm_MUVR-QA_results} and have the following findings: 

\textbf{Finding 1: Tag prompts pose significant challenges for MLLMs.} While some MLLMs achieve above 70\% accuracy on questions without tag prompts (No Tag), performance drops substantially when handling questions incorporating tag prompts (All). 

\textbf{Finding 2: Multi-frame processing improves performance but increases latency.} Processing multiple frames (6-12 frames) significantly enhances both Accuracy and Reranking Score compared to single-frame input (e.g., InternVL2.5 improves from 52.0\% to 57.0\% Accuracy), though at the cost of higher inference time (from 1.49s to 4.23s). However, beyond a certain point (12 frames), performance gains diminish while computational overhead continues to rise. 

\textbf{Finding 3: Multi-image and mask prompt understanding capabilities boost reranking effectiveness.} Models with joint query-target video frame processing (e.g., InternVL, MiniCPM) consistently outperform text query-only comparison methods in the Reranking Score metric, demonstrating the value of direct visual comparison. Besides, the recent VideoRefer can understand the mask prompt and achieves better performance with the mask prompt. 

More analysis is available in the Appendix.

\subsection{Robustness Analysis}
\label{sec:annotation_noise}

To evaluate the robustness of MUVR against annotation noise, we conducted experiments simulating false negatives and annotation errors. Specifically, we randomly increased or decreased the number of positive samples by 5\% relative to the total number of positive samples, which was repeated five times. As shown in Table~\ref{tab:positive_samples_results}, increasing the number of positive samples leads to a slight improvement in performance compared to Table~\ref{tab: MUVR-Base combined results}(refer to "mAP of different partitions", "multimodal query"), while decreasing the number of positive samples results in a more noticeable drop in performance. This highlights the importance of accurate annotations. Importantly, the relative ranking of methods remains highly stable across these experiments, demonstrating the robustness of our benchmark.

\begin{table*}[!t]
\centering
\caption{Performance comparison with +5\% and -5\% positive samples.}
\label{tab:positive_samples_results}
\small
\setlength{\tabcolsep}{2mm}
\begin{tabular}{lccccc}
\toprule
\textbf{Method} & \textbf{News (N)} & \textbf{Others (O)} & \textbf{Instance (I)} & \textbf{Region (R)} & \textbf{Dance (D)} \\
\midrule
\rowcolor{blue!10}
\multicolumn{6}{c}{\textbf{\textit{+5\% Positive Samples}}} \\
CLIP (RN50x4)\cite{clip} & 49.6±0.1 & 53.8±0.1 & 46.8±0.1 & 44.1±0.1 & 21.5±0.1 \\
CLIP (ViT-L/14@336px)\cite{clip} & 58.4±0.1 & 57.9±0.1 & 54.4±0.1 & 51.0±0.1 & 25.5±0.1 \\
OpenCLIP (ViT-H-14)\cite{openclip} & 59.6±0.1 & 62.8±0.1 & 62.6±0.2 & 59.3±0.1 & 26.8±0.2 \\
EVA-CLIP\cite{evaclip} & 63.4±0.1 & 66.2±0.1 & 68.4±0.1 & 64.1±0.1 & 29.1±0.1 \\
BLIP\cite{blip} & 50.5±0.1 & 54.4±0.1 & 49.6±0.1 & 47.2±0.1 & 19.8±0.1 \\
BLIP2\cite{blip2} & 56.5±0.1 & 61.8±0.1 & 58.3±0.1 & 54.2±0.2 & 25.7±0.1 \\
\midrule
\rowcolor{blue!10}
\multicolumn{6}{c}{\textbf{\textit{-5\% Positive Samples}}} \\
CLIP (RN50x4)\cite{clip} & 43.1±0.3 & 47.7±0.2 & 41.8±0.4 & 39.3±0.4 & 19.1±0.1 \\
CLIP (ViT-L/14@336px)\cite{clip} & 51.6±0.6 & 51.7±0.1 & 48.6±0.2 & 45.1±0.1 & 22.7±0.2 \\
OpenCLIP (ViT-H-14)\cite{openclip} & 52.6±0.1 & 56.1±0.3 & 55.6±0.1 & 52.6±0.1 & 24.0±0.2 \\
EVA-CLIP\cite{evaclip} & 56.0±0.4 & 59.4±0.5 & 60.8±0.5 & 57.0±0.1 & 25.7±0.3 \\
BLIP\cite{blip} & 44.6±0.4 & 48.4±0.1 & 44.1±0.1 & 42.0±0.1 & 17.5±0.2 \\
BLIP2\cite{blip2} & 49.7±0.1 & 54.9±0.1 & 51.8±0.5 & 47.8±0.2 & 22.6±0.1 \\
\bottomrule
\end{tabular}
\end{table*}

\section{Conclusion and Future Work}
\label{sec:conclusion and Future Work}
This paper introduces the Multi-modal Untrimmed Video Retrieval task and benchmark (MUVR) to address the limitations of existing video retrieval tasks in handling untrimmed videos and diverse query modalities. MUVR features a practical retrieval paradigm supporting video-centric multi-modal queries, organizes videos into five partitions based on multi-level visual correspondence, and provides comprehensive evaluation protocols including a novel Reranking Score for assessing MLLMs. Experimental results reveal significant challenges in the current models' ability to process untrimmed videos and multi-modal queries, as well as MLLMs' limitations in multi-video understanding. 

\textbf{Future work} should focus on developing more effective fusion methods for multimodal queries, improving temporal modeling for long videos, and enhancing MLLMs' efficiency and multiple video understanding capabilities for better reranking performance. The benchmark's diverse video categories and flexible query formats offer rich opportunities for advancing video retrieval research.

\section*{Acknowledgements} 
This work was partially supported by the National Natural Science Foundation of China (No. 62276129),  the Natural Science Foundation of Jiangsu Province (No. BK20250082), the Fundamental Research Funds for the Central Universities (No. NE2025010) and the Jiangsu Funding Program for Excellent Postdoctoral Talent (No. 2025ZB306).

% \section*{References}

% \medskip

{
\small
\bibliographystyle{IEEEtran}
\bibliography{neurips_2025}
}

%%%%%%%%%%%%%%%%%%%%%%%%%%%%%%%%%%%%%%%%%%%%%%%%%%%%%%%%%%%%

% \appendix

% \input{sections/appendix}

%%%%%%%%%%%%%%%%%%%%%%%%%%%%%%%%%%%%%%%%%%%%%%%%%%%%%%%%%%%%

\newpage
\section*{NeurIPS Paper Checklist}

\begin{enumerate}

\item {\bf Claims}
    \item[] Question: Do the main claims made in the abstract and introduction accurately reflect the paper's contributions and scope?
    \item[] Answer: \answerYes{} % Replace by \answerYes{}, \answerNo{}, or \answerNA{}.
    \item[] Justification: We gave our claim and contributions in Section \ref{sec:intro}.
    % \item[] Guidelines:
    % \begin{itemize}
    %     \item The answer NA means that the abstract and introduction do not include the claims made in the paper.
    %     \item The abstract and/or introduction should clearly state the claims made, including the contributions made in the paper and important assumptions and limitations. A No or NA answer to this question will not be perceived well by the reviewers. 
    %     \item The claims made should match theoretical and experimental results, and reflect how much the results can be expected to generalize to other settings. 
    %     \item It is fine to include aspirational goals as motivation as long as it is clear that these goals are not attained by the paper. 
    % \end{itemize}

\item {\bf Limitations}
    \item[] Question: Does the paper discuss the limitations of the work performed by the authors?
    \item[] Answer: \answerYes{} % Replace by \answerYes{}, \answerNo{}, or \answerNA{}.
    \item[] Justification: We discussed the limitations of our article by looking ahead to the future research directions in Section \ref{sec:conclusion and Future Work}.

\item {\bf Theory assumptions and proofs}
    \item[] Question: For each theoretical result, does the paper provide the full set of assumptions and a complete (and correct) proof?
    \item[] Answer: \answerNA{} % Replace by \answerYes{}, \answerNo{}, or \answerNA{}.
    \item[] Justification: This is not a theoretical paper.
    % \item[] Guidelines:
    % \begin{itemize}
    %     \item The answer NA means that the paper does not include theoretical results. 
    %     \item All the theorems, formulas, and proofs in the paper should be numbered and cross-referenced.
    %     \item All assumptions should be clearly stated or referenced in the statement of any theorems.
    %     \item The proofs can either appear in the main paper or the supplemental material, but if they appear in the supplemental material, the authors are encouraged to provide a short proof sketch to provide intuition. 
    %     \item Inversely, any informal proof provided in the core of the paper should be complemented by formal proofs provided in appendix or supplemental material.
    %     \item Theorems and Lemmas that the proof relies upon should be properly referenced. 
    % \end{itemize}
    
    \item {\bf Experimental result reproducibility}
    \item[] Question: Does the paper fully disclose all the information needed to reproduce the main experimental results of the paper to the extent that it affects the main claims and/or conclusions of the paper (regardless of whether the code and data are provided or not)?
    \item[] Answer: \answerYes{} % Replace by \answerYes{}, \answerNo{}, or \answerNA{}.
    \item[] Justification: We didn't propose a model ourselves. Instead, we used someone else's model and referred to their hyperparameters.

\item {\bf Open access to data and code}
    \item[] Question: Does the paper provide open access to the data and code, with sufficient instructions to faithfully reproduce the main experimental results, as described in supplemental material?
    \item[] Answer: \answerYes{} % Replace by \answerYes{}, \answerNo{}, or \answerNA{}.
    \item[] Justification: We have released our benchmark and our code.
    % \item[] Guidelines:
    % \begin{itemize}
    %     \item The answer NA means that paper does not include experiments requiring code.
    %     \item Please see the NeurIPS code and data submission guidelines (\url{https://nips.cc/public/guides/CodeSubmissionPolicy}) for more details.
    %     \item While we encourage the release of code and data, we understand that this might not be possible, so “No” is an acceptable answer. Papers cannot be rejected simply for not including code, unless this is central to the contribution (e.g., for a new open-source benchmark).
    %     \item The instructions should contain the exact command and environment needed to run to reproduce the results. See the NeurIPS code and data submission guidelines (\url{https://nips.cc/public/guides/CodeSubmissionPolicy}) for more details.
    %     \item The authors should provide instructions on data access and preparation, including how to access the raw data, preprocessed data, intermediate data, and generated data, etc.
    %     \item The authors should provide scripts to reproduce all experimental results for the new proposed method and baselines. If only a subset of experiments are reproducible, they should state which ones are omitted from the script and why.
    %     \item At submission time, to preserve anonymity, the authors should release anonymized versions (if applicable).
    %     \item Providing as much information as possible in supplemental material (appended to the paper) is recommended, but including URLs to data and code is permitted.
    % \end{itemize}

\item {\bf Experimental setting/details}
    \item[] Question: Does the paper specify all the training and test details (e.g., data splits, hyperparameters, how they were chosen, type of optimizer, etc.) necessary to understand the results?
    \item[] Answer: \answerYes{} % Replace by \answerYes{}, \answerNo{}, or \answerNA{}.
    \item[] Justification: The details are in Section \ref{sec: Models and Methods for Evaluation} and Appendix.
    % \item[] Guidelines:
    % \begin{itemize}
    %     \item The answer NA means that the paper does not include experiments.
    %     \item The experimental setting should be presented in the core of the paper to a level of detail that is necessary to appreciate the results and make sense of them.
    %     \item The full details can be provided either with the code, in appendix, or as supplemental material.
    % \end{itemize}

\item {\bf Experiment statistical significance}
    \item[] Question: Does the paper report error bars suitably and correctly defined or other appropriate information about the statistical significance of the experiments?
    \item[] Answer: \answerNo{} % Replace by \answerYes{}, \answerNo{}, or \answerNA{}.
    \item[] Justification: We do not report error bars.
    % \item[] Guidelines:
    % \begin{itemize}
    %     \item The answer NA means that the paper does not include experiments.
    %     \item The authors should answer "Yes" if the results are accompanied by error bars, confidence intervals, or statistical significance tests, at least for the experiments that support the main claims of the paper.
    %     \item The factors of variability that the error bars are capturing should be clearly stated (for example, train/test split, initialization, random drawing of some parameter, or overall run with given experimental conditions).
    %     \item The method for calculating the error bars should be explained (closed form formula, call to a library function, bootstrap, etc.)
    %     \item The assumptions made should be given (e.g., Normally distributed errors).
    %     \item It should be clear whether the error bar is the standard deviation or the standard error of the mean.
    %     \item It is OK to report 1-sigma error bars, but one should state it. The authors should preferably report a 2-sigma error bar than state that they have a 96\% CI, if the hypothesis of Normality of errors is not verified.
    %     \item For asymmetric distributions, the authors should be careful not to show in tables or figures symmetric error bars that would yield results that are out of range (e.g. negative error rates).
    %     \item If error bars are reported in tables or plots, The authors should explain in the text how they were calculated and reference the corresponding figures or tables in the text.
    % \end{itemize}

\item {\bf Experiments compute resources}
    \item[] Question: For each experiment, does the paper provide sufficient information on the computer resources (type of compute workers, memory, time of execution) needed to reproduce the experiments?
    \item[] Answer: \answerYes{} % Replace by \answerYes{}, \answerNo{}, or \answerNA{}.
    \item[] Justification: We provide the information in Section \ref{sec: experiments}.
    % \item[] Guidelines:
    % \begin{itemize}
    %     \item The answer NA means that the paper does not include experiments.
    %     \item The paper should indicate the type of compute workers CPU or GPU, internal cluster, or cloud provider, including relevant memory and storage.
    %     \item The paper should provide the amount of compute required for each of the individual experimental runs as well as estimate the total compute. 
    %     \item The paper should disclose whether the full research project required more compute than the experiments reported in the paper (e.g., preliminary or failed experiments that didn't make it into the paper). 
    % \end{itemize}
    
\item {\bf Code of ethics}
    \item[] Question: Does the research conducted in the paper conform, in every respect, with the NeurIPS Code of Ethics \url{https://neurips.cc/public/EthicsGuidelines}?
    \item[] Answer: \answerYes{} % Replace by \answerYes{}, \answerNo{}, or \answerNA{}.
    \item[] Justification: We have read and understood the code of ethics and have done our best to conform.
    % \item[] Guidelines:
    % \begin{itemize}
    %     \item The answer NA means that the authors have not reviewed the NeurIPS Code of Ethics.
    %     \item If the authors answer No, they should explain the special circumstances that require a deviation from the Code of Ethics.
    %     \item The authors should make sure to preserve anonymity (e.g., if there is a special consideration due to laws or regulations in their jurisdiction).
    % \end{itemize}

\item {\bf Broader impacts}
    \item[] Question: Does the paper discuss both potential positive societal impacts and negative societal impacts of the work performed?
    \item[] Answer: \answerYes{} % Replace by \answerYes{}, \answerNo{}, or \answerNA{}.
    \item[] Justification: We analyzed the societal impacts of our work in Section\ref{sec:intro} and Appendix~\ref{sec4}

\item {\bf Safeguards}
    \item[] Question: Does the paper describe safeguards that have been put in place for responsible release of data or models that have a high risk for misuse (e.g., pretrained language models, image generators, or scraped datasets)?
    \item[] Answer: \answerNA{} % Replace by \answerYes{}, \answerNo{}, or \answerNA{}.
    \item[] Justification: The paper poses no such risks.
    % \item[] Guidelines:
    % \begin{itemize}
    %     \item The answer NA means that the paper poses no such risks.
    %     \item Released models that have a high risk for misuse or dual-use should be released with necessary safeguards to allow for controlled use of the model, for example by requiring that users adhere to usage guidelines or restrictions to access the model or implementing safety filters. 
    %     \item Datasets that have been scraped from the Internet could pose safety risks. The authors should describe how they avoided releasing unsafe images.
    %     \item We recognize that providing effective safeguards is challenging, and many papers do not require this, but we encourage authors to take this into account and make a best faith effort.
    % \end{itemize}

\item {\bf Licenses for existing assets}
    \item[] Question: Are the creators or original owners of assets (e.g., code, data, models), used in the paper, properly credited and are the license and terms of use explicitly mentioned and properly respected?
    \item[] Answer: \answerYes{} % Replace by \answerYes{}, \answerNo{}, or \answerNA{}.
    \item[] Justification: We credited the authors of the models considered and code we used in appropriate ways.
    % \item[] Guidelines:
    % \begin{itemize}
    %     \item The answer NA means that the paper does not use existing assets.
    %     \item The authors should cite the original paper that produced the code package or dataset.
    %     \item The authors should state which version of the asset is used and, if possible, include a URL.
    %     \item The name of the license (e.g., CC-BY 4.0) should be included for each asset.
    %     \item For scraped data from a particular source (e.g., website), the copyright and terms of service of that source should be provided.
    %     \item If assets are released, the license, copyright information, and terms of use in the package should be provided. For popular datasets, \url{paperswithcode.com/datasets} has curated licenses for some datasets. Their licensing guide can help determine the license of a dataset.
    %     \item For existing datasets that are re-packaged, both the original license and the license of the derived asset (if it has changed) should be provided.
    %     \item If this information is not available online, the authors are encouraged to reach out to the asset's creators.
    % \end{itemize}

\item {\bf New assets}
    \item[] Question: Are new assets introduced in the paper well documented and is the documentation provided alongside the assets?
    \item[] Answer: \answerYes{} % Replace by \answerYes{}, \answerNo{}, or \answerNA{}.
    \item[] Justification: We have released our benchmark and provided a detailed README.md document. Also, please see Section \ref{sec:MUVR Benchmark} for dataset construction details.

    % \item[] Guidelines:
    % \begin{itemize}
    %     \item The answer NA means that the paper does not release new assets.
    %     \item Researchers should communicate the details of the dataset/code/model as part of their submissions via structured templates. This includes details about training, license, limitations, etc. 
    %     \item The paper should discuss whether and how consent was obtained from people whose asset is used.
    %     \item At submission time, remember to anonymize your assets (if applicable). You can either create an anonymized URL or include an anonymized zip file.
    % \end{itemize}

\item {\bf Crowdsourcing and research with human subjects}
    \item[] Question: For crowdsourcing experiments and research with human subjects, does the paper include the full text of instructions given to participants and screenshots, if applicable, as well as details about compensation (if any)? 
    \item[] Answer: \answerYes{} % Replace by \answerYes{}, \answerNo{}, or \answerNA{}.
    \item[] Justification: Please see Section \ref{sec:Composition and Collection} for details.
    % \item[] Guidelines:
    % \begin{itemize}
    %     \item The answer NA means that the paper does not involve crowdsourcing nor research with human subjects.
    %     \item Including this information in the supplemental material is fine, but if the main contribution of the paper involves human subjects, then as much detail as possible should be included in the main paper. 
    %     \item According to the NeurIPS Code of Ethics, workers involved in data collection, curation, or other labor should be paid at least the minimum wage in the country of the data collector. 
    % \end{itemize}

\item {\bf Institutional review board (IRB) approvals or equivalent for research with human subjects}
    \item[] Question: Does the paper describe potential risks incurred by study participants, whether such risks were disclosed to the subjects, and whether Institutional Review Board (IRB) approvals (or an equivalent approval/review based on the requirements of your country or institution) were obtained?
    \item[] Answer: \answerYes{} % Replace by \answerYes{}, \answerNo{}, or \answerNA{}.
    \item[] Justification: Please see Section \ref{sec:Composition and Collection} for details.
    % \item[] Guidelines:
    % \begin{itemize}
    %     \item The answer NA means that the paper does not involve crowdsourcing nor research with human subjects.
    %     \item Depending on the country in which research is conducted, IRB approval (or equivalent) may be required for any human subjects research. If you obtained IRB approval, you should clearly state this in the paper. 
    %     \item We recognize that the procedures for this may vary significantly between institutions and locations, and we expect authors to adhere to the NeurIPS Code of Ethics and the guidelines for their institution. 
    %     \item For initial submissions, do not include any information that would break anonymity (if applicable), such as the institution conducting the review.
    % \end{itemize}

\item {\bf Declaration of LLM usage}
    \item[] Question: Does the paper describe the usage of LLMs if it is an important, original, or non-standard component of the core methods in this research? Note that if the LLM is used only for writing, editing, or formatting purposes and does not impact the core methodology, scientific rigorousness, or originality of the research, declaration is not required.
    %this research? 
    \item[] Answer: \answerNA{} % Replace by \answerYes{}, \answerNo{}, or \answerNA{}.
    \item[] Justification:  The LLM is used only for writing, editing, or formatting purposes in our paper.
    % \item[] Guidelines:
    % \begin{itemize}
    %     \item The answer NA means that the core method development in this research does not involve LLMs as any important, original, or non-standard components.
    %     \item Please refer to our LLM policy (\url{https://neurips.cc/Conferences/2025/LLM}) for what should or should not be described.
    % \end{itemize}

\end{enumerate}

\clearpage
\section*{Appendix}
\appendix

In this Appendix, more analysis of results is provided in Section \ref{sec1}. Limitations and social impact are introduced in Section \ref{sec2}. we further elaborate on the MLLMs prompting details in Section \ref{sec3}. We further illustrate the annotation instructions in Section \ref{sec4}. Finally, some visualization examples are provided in Section \ref{sec5}.

\section{More Analysis of Results}
\label{sec1}
\subsection{Retrieval with Video and Text Queries}

\textbf{Difficulty in Aligning Fine-Grained Instance-Level Visual Information:} We observed a significant performance gap between video-only queries and text-only queries in instance-level partitions (e.g., CLIP RN50x4: video 31.8\% vs. text 41.0\%). This discrepancy arises because video queries often introduce excessive background noise, whereas instance-level partitions focus on small objects such as products or pets. Although embedding models can align instance-level features in the vision-language space, their visual embeddings struggle to disentangle foreground and background information, making fine-grained visual alignment challenging.

\textbf{Weak Temporal Modeling for Dynamic Actions:} All models performed the worst in the dance partition (e.g., even video-based VLMs like InternVideo2 achieved only 28.9\%). This highlights the limitations of current temporal modeling strategies in capturing the complex, fast, and prolonged motion patterns of dance sequences, which are even more intricate than sports activities like diving. This finding provides valuable insights for future research on improving temporal modeling capabilities.

\subsection{Retrieval with Additional Tag Prompts}

\textbf{Limited Ability to Integrate Tag Prompts with Queries:} Embedding models showed only marginal performance improvements when tag prompts were added. This is primarily due to the lack of effective methods for injecting tag semantics into the query. Additionally, the tags themselves often contain challenging semantic information, such as video styles or perspectives. Future work could explore leveraging large language models (LLMs) to understand better and integrate these tags.

\textbf{Inability to Handle Complex Tag Combinations:} Current evaluations are limited to single positive or negative tag prompts. Ideally, embedding models should be capable of filtering retrieval results by incorporating multiple positive and negative tags, similar to how text-to-image generation models process prompts with complex tag combinations. We plan to systematically explore retrieval frameworks that integrate semantic tags more effectively in future work.

\subsection{Reranking Performance of MLLMs}

\textbf{Low Efficiency in Retrieval/Reranking with Large Models:} While MLLMs have the potential to outperform embedding models in retrieval tasks (e.g., multimodal query understanding and relevance assessment), their current approach to comparing two videos introduces significant inference delays. A feasible solution is to reduce the number of visual tokens during the final decision-making stage using token compression techniques.

\textbf{Lack of Multi-Video Understanding Capabilities:} To the best of our knowledge, most existing MLLMs are not optimized for multi-video understanding, making it difficult to assess the relevance between two input videos. One potential solution is to insert separator tokens between the input videos to help the model distinguish their sources, followed by LoRA fine-tuning to enhance performance.

\textbf{Need for Improved Fine-Grained Spatiotemporal Understanding:} We observed that models like VideoRefer, which support masked inputs, can better understand the specific instances users aim to retrieve, thereby improving reranking performance. This suggests that enhancing MLLMs' fine-grained spatiotemporal understanding could enable them to capture the key features of the query more accurately. Future work will focus on this direction to further improve reranking capabilities.

\section{Limitations and Social Impact}
\label{sec2}
\textbf{Limitations.}
MUVR relies on human annotators to annotate videos with rich semantics. Despite strict guidance for annotators and multiple rounds of validation during the annotation process, there may still be minor annotation errors. Besides, MUVR focuses on visual and textual modalities, leaving out other potential modalities such as audio, which could further enrich the retrieval task. Despite these limitations, we believe MUVR offers a robust foundation for advancing research in video retrieval, and its design allows for future extensions to address these gaps.

\textbf{Social Impact.}
The development of MUVR has potential positive implications for improving video search and recommendation systems, enhancing user experience on video-sharing platforms. By enabling more accurate and fine-grained retrieval, our work could facilitate better access to educational, informational, and entertainment content.

\textbf{Ethical Considerations.}
All videos used in MUVR were downloaded in strict adherence to the copyright and terms of service of the respective platforms, solely for scientific research purposes. To ensure transparency and reproducibility, we release the video IDs, annotation files, and pre-extracted features. Researchers can access the videos directly from the platforms, provided they comply with licensing terms. Additionally, a takedown mechanism is available on our project website, allowing copyright holders to request removal of their content. We believe this approach aligns with ethical standards and copyright laws, ensuring responsible use of publicly available data for research purposes.

% 单阶段文本查询模型
\begin{table}[h]
\centering
\caption{Format of the text prompts used by MLLMs for one-stage text query-only comparison. \textcolor{blue}{<Target Video>}: format as `Frame1: <image>\textbackslash nFrame2: <image>\textbackslash n...Frame6: <image>\textbackslash n'.}
\label{tab: prompt one stage}
\small
\begin{tabular}{l p{0.7\textwidth}}
\toprule
\textbf{Model} & \textbf{Text Prompt} \\
\midrule
InternVL2\cite{wang2024internvideo2scalingfoundationmodels} & \fontsize{9}{11}\selectfont I will give you a text query and a video: [Query] and [Target]. Please determine whether any part of [Target] is slightly relevant to any part of [Query]. I will also provide [Tag] that [Target] (if relevant) must feature it.\textbackslash n[Query]:\textbackslash n\textcolor{blue}{\{Text Description\}}\textbackslash n[Target]:\textbackslash n\textcolor{blue}{<Target Video>}\textbackslash n[Tag]:\textbackslash n\textcolor{blue}{\{Tag Prompt\}}\textbackslash n[Output]:\textbackslash n If slightly relevant, return Yes. If not, return No. \\ 
\midrule
InternVL2.5\cite{internvl25} & \fontsize{9}{11}\selectfont I will give you a text query and a video: [Query] and [Target]. Please determine whether any part of [Target] is slightly relevant to any part of [Query]. I will also provide [Tag] that [Target] (if relevant) must feature it.\textbackslash n[Query]:\textbackslash n\textcolor{blue}{\{Text Description\}}\textbackslash n[Target]:\textbackslash n\textcolor{blue}{<Target Video>}\textbackslash n[Tag]:\textbackslash n\textcolor{blue}{\{Tag Prompt\}}\textbackslash n[Output]:\textbackslash n If slightly relevant, return Yes. If not, return No. \\
\midrule
MiniCPM-o 2.6\cite{minicpm} & \fontsize{9}{11}\selectfont Please determine whether any part of the video is slightly relevant to any part of [Query]. I will also provide [Tag] that the video (if relevant) must feature it. [Query]: \textcolor{blue}{\{Text Description\}}\textbackslash n[Tag]: \textcolor{blue}{\{Tag Prompt\}}\textbackslash n If slightly relevant, return Yes. If not, return No. \\
\midrule
MiniCPM-V 2.6\cite{minicmpv} & \fontsize{9}{11}\selectfont Please determine whether any part of the video is slightly relevant to any part of [Query]. I will also provide [Tag] that the video (if relevant) must feature it. [Query]: \textcolor{blue}{\{Text Description\}}\textbackslash n[Tag]: \textcolor{blue}{\{Tag Prompt\}}\textbackslash n If slightly relevant, return Yes. If not, return No. \\
\midrule
LLaVA-NeXT-Video\cite{llavanext} & \fontsize{9}{11}\selectfont Please determine whether any part of the video is slightly relevant to any part of [Query]. I will also provide [Tag] that the video (if relevant) must feature it. [Query]: \textcolor{blue}{\{Text Description\}}\textbackslash n[Tag]: \textcolor{blue}{\{Tag Prompt\}}\textbackslash n If slightly relevant, return Yes. If not, return No. \\
\midrule
LLaVA-OV\cite{llavaonevision} & \fontsize{9}{11}\selectfont Please determine whether any part of the video is slightly relevant to any part of [Query]. I will also provide [Tag] that the video (if relevant) must feature it. [Query]: \textcolor{blue}{\{Text Description\}}\textbackslash n[Tag]: \textcolor{blue}{\{Tag Prompt\}}\textbackslash n If slightly relevant, return Yes. If not, return No. \\
\midrule
LLaVA-Video\cite{llavavideo} & \fontsize{9}{11}\selectfont Please determine whether any part of the video is slightly relevant to any part of [Query]. I will also provide [Tag] that the video (if relevant) must feature it. [Query]: \textcolor{blue}{\{Text Description\}}\textbackslash n[Tag]: \textcolor{blue}{\{Tag Prompt\}}\textbackslash n If slightly relevant, return Yes. If not, return No. \\
\bottomrule
\end{tabular}
\end{table}

\section{MLLMs Prompting Details}
\label{sec3}
The evaluation prompts for MLLMs are listed in Table \ref{tab: prompt one stage} and \ref{tab: prompt multi image}. Although we attempted to maintain consistency across models, slight variations were necessary due to differing prompting requirements. The proprietary models (GPT-4o and Gemini-2.0-Flash) were accessed on April 25, 2025.

% 多图理解模型
\begin{table}[h]
\centering
\caption{Format of the text prompts used by MLLMs for one-stage multi-image comparison. \textcolor{blue}{<Query Video>}/\textcolor{blue}{<Target Video>}: format as `Frame1: <image>\textbackslash nFrame2: <image>\textbackslash n...Frame6: <image>\textbackslash n'. $\dagger$: using additional mask prompt.}
\label{tab: prompt multi image}
\small
\begin{tabular}{l p{0.7\textwidth}}
\toprule
\textbf{Model} & \textbf{Text Prompt} \\
\midrule 
InternVL2\cite{wang2024internvideo2scalingfoundationmodels} & \fontsize{9}{11}\selectfont I will give you a video query and a video target: [Query] and [Target]. Please determine whether any part of [Target] is slightly relevant to any part of [Query] or [Focus]. I will also provide [Tag] that [Target] (if relevant) must feature it.\textbackslash n[Query]:\textbackslash n\textcolor{blue}{<Query Video>}\textbackslash n[Target]:\textbackslash n\textcolor{blue}{<Target Video>}\textbackslash n[Focus]\textbackslash n\textcolor{blue}{\{Text Description\}}\textbackslash n[Tag]:\textbackslash n\textcolor{blue}{\{Tag Prompt\}}\textbackslash n[Output]:\textbackslash n If slightly relevant, return Yes. If not, return No. \\ 
\midrule
InternVL2.5\cite{internvl25} & \fontsize{9}{11}\selectfont I will give you a video query and a video target: [Query] and [Target]. Please determine whether any part of [Target] is slightly relevant to any part of [Query] or [Focus]. I will also provide [Tag] that [Target] (if relevant) must feature it.\textbackslash n[Query]:\textbackslash n\textcolor{blue}{<Query Video>}\textbackslash n[Target]:\textbackslash n\textcolor{blue}{<Target Video>}\textbackslash n[Focus]\textbackslash n\textcolor{blue}{\{Text Description\}}\textbackslash n[Tag]:\textbackslash n\textcolor{blue}{\{Tag Prompt\}}\textbackslash n[Output]:\textbackslash n If slightly relevant, return Yes. If not, return No. \\
\midrule
MiniCPM-o 2.6\cite{minicpm} & \fontsize{9}{11}\selectfont Please determine whether any part of \textcolor{blue}{<Target Video>} is slightly relevant to any part of \textcolor{blue}{<Query Video>} and [Focus]. I will also provide [Tag] that \textcolor{blue}{<Target Video>} (if relevant) must feature it. [Focus]: \textcolor{blue}{\{Text Description\}}\textbackslash n[Tag]: \textcolor{blue}{\{Tag Prompt\}}\textbackslash n If slightly relevant, return Yes. If not, return No. \\
\midrule
MiniCPM-V 2.6\cite{minicmpv} & \fontsize{9}{11}\selectfont Please determine whether any part of \textcolor{blue}{<Target Video>} is slightly relevant to any part of \textcolor{blue}{<Query Video>} and [Focus]. I will also provide [Tag] that \textcolor{blue}{<Target Video>} (if relevant) must feature it. [Focus]: \textcolor{blue}{\{Text Description\}}\textbackslash n[Tag]: \textcolor{blue}{\{Tag Prompt\}}\textbackslash n If slightly relevant, return Yes. If not, return No. \\
\midrule
VideoRefer\cite{yuan2025videorefersuite} & \fontsize{9}{11}\selectfont Here are two videos with same length. Is any part of the first video query slightly relevant to any part of the second video? \textcolor{blue}{\{Text Description\}}\textbackslash n If true and \textcolor{blue}{\{Tag Prompt\}}, return Yes. Else, return No. \\
\midrule
VideoRefer$\dagger$\cite{yuan2025videorefersuite} & Here are two videos with same length. Is any part of the first video query slightly relevant to any part of the second video? \textcolor{blue}{\{Text Description\}}\textbackslash n If true and \textcolor{blue}{\{Tag Prompt\}}, return Yes. Else, return No. \\
\midrule
Gemini-2.0-Flash\cite{gemini} & \fontsize{9}{11}\selectfont Is any part of the first video query slightly relevant to any part of the second video? \textcolor{blue}{\{Text Description\}}\textbackslash n If true and \textcolor{blue}{\{Tag Prompt\}}, return Yes. Else, return No. \\
\midrule
GPT-4o\cite{openai2024gpt4o} & \fontsize{9}{11}\selectfont Here are two videos with same length. Is any part of the first video query slightly relevant to any part of the second video? \textcolor{blue}{\{Text Description\}}\textbackslash n If true and \textcolor{blue}{\{Tag Prompt\}}, return Yes. Else, return No. \\

\bottomrule
\end{tabular}
\end{table}

\section{Annotation Instructions}
\label{sec4}
The instructions provided to annotators are included below. We take the relationship annotation of the News partition as an example, while other partitions have different visual correspondences.

\begin{figure}[h] % [H] 强制图片位置
    \centering
    \setlength{\fboxsep}{0pt} % 图片与边框间距
    \setlength{\fboxrule}{1pt} % 边框粗细
    
    % 第一行
    \fbox{\includegraphics[width=0.49\textwidth]{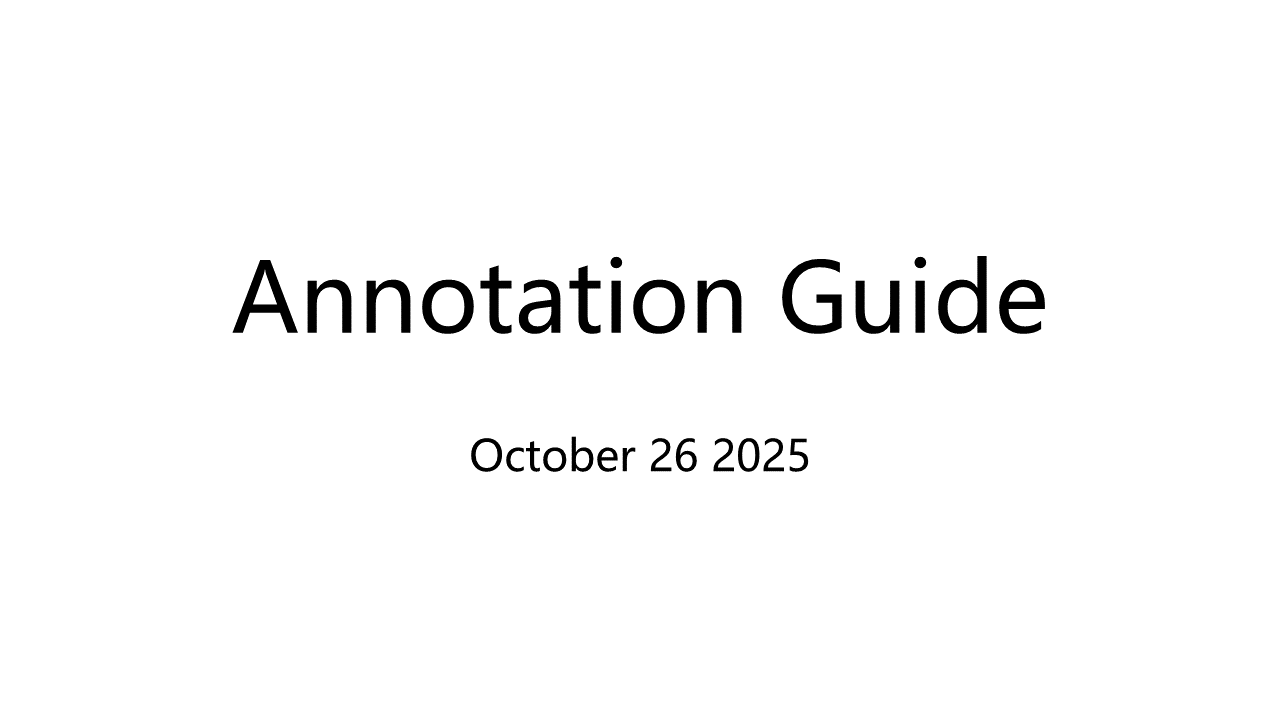}}%
    \hspace{0.0\textwidth}% 调整图片间距（可选）
    \fbox{\includegraphics[width=0.49\textwidth]{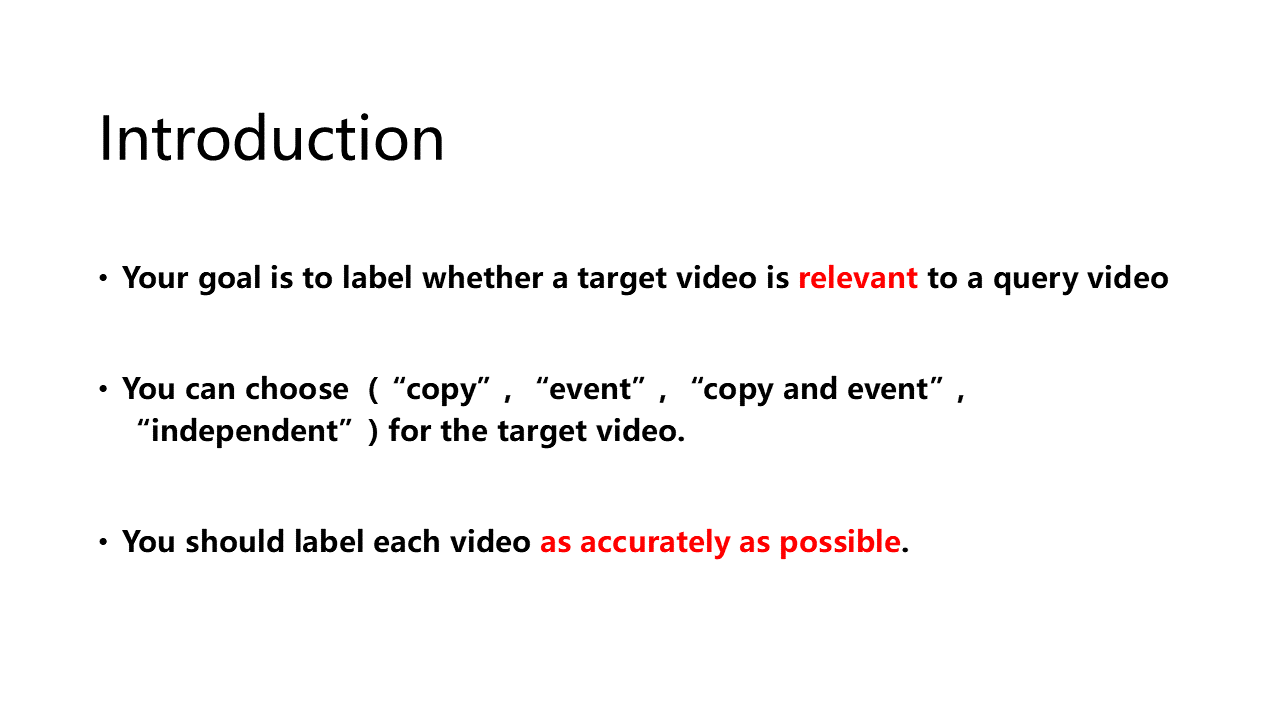}}%
    
    % 第二行
    \vspace{-0.1em}% 调整行间距（可选）
    \fbox{\includegraphics[width=0.49\textwidth]{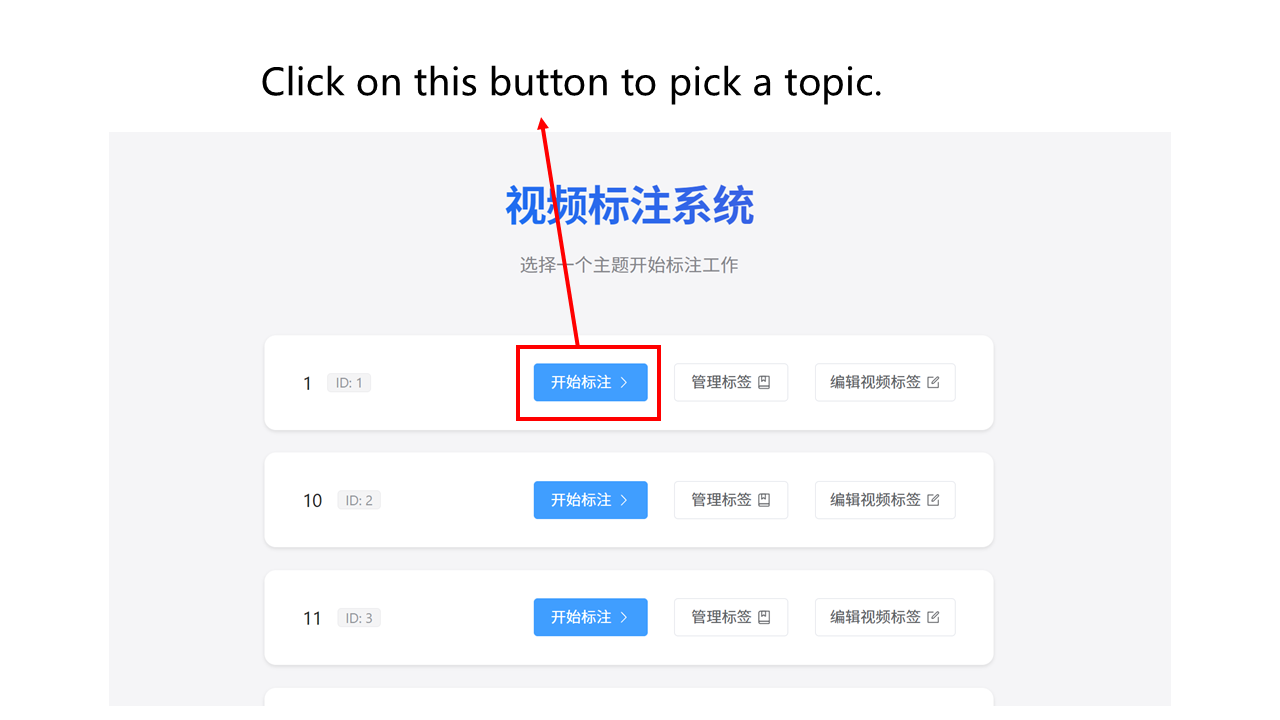}}%
    \hspace{0.0\textwidth}%
    \fbox{\includegraphics[width=0.49\textwidth]{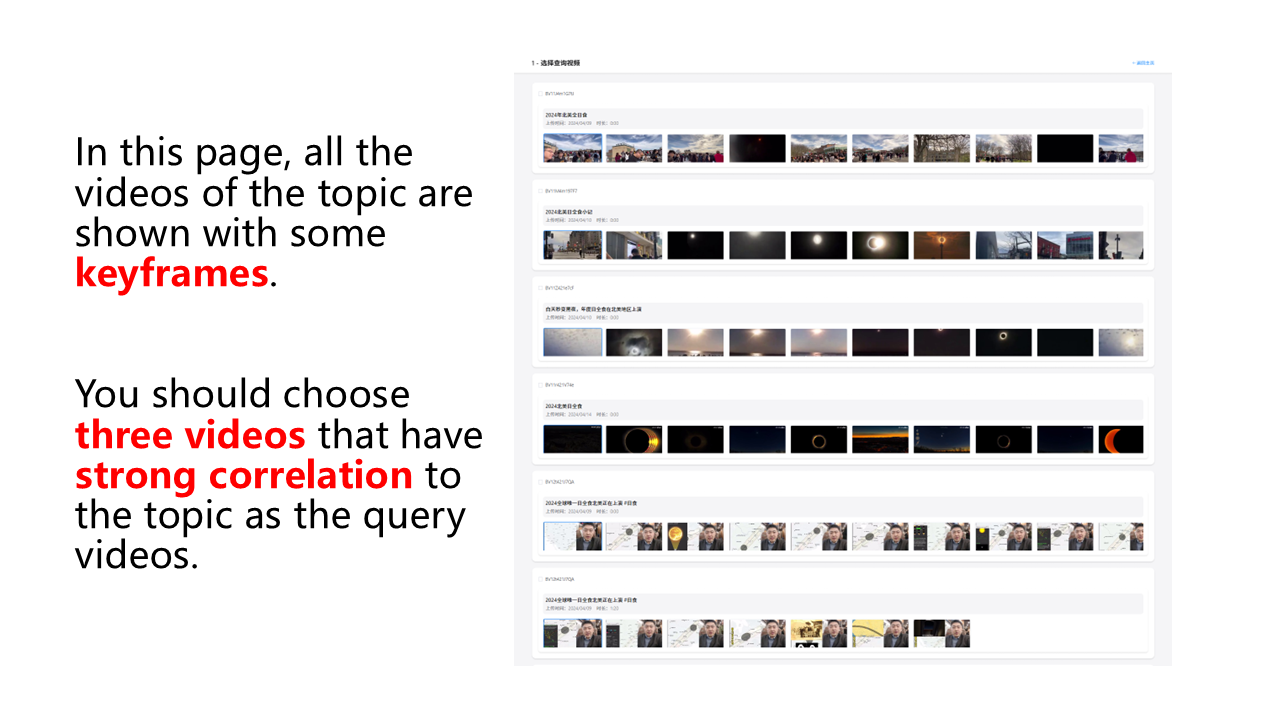}}%
    
    % 第三行
    \vspace{-0.1em}%
    \fbox{\includegraphics[width=0.49\textwidth]{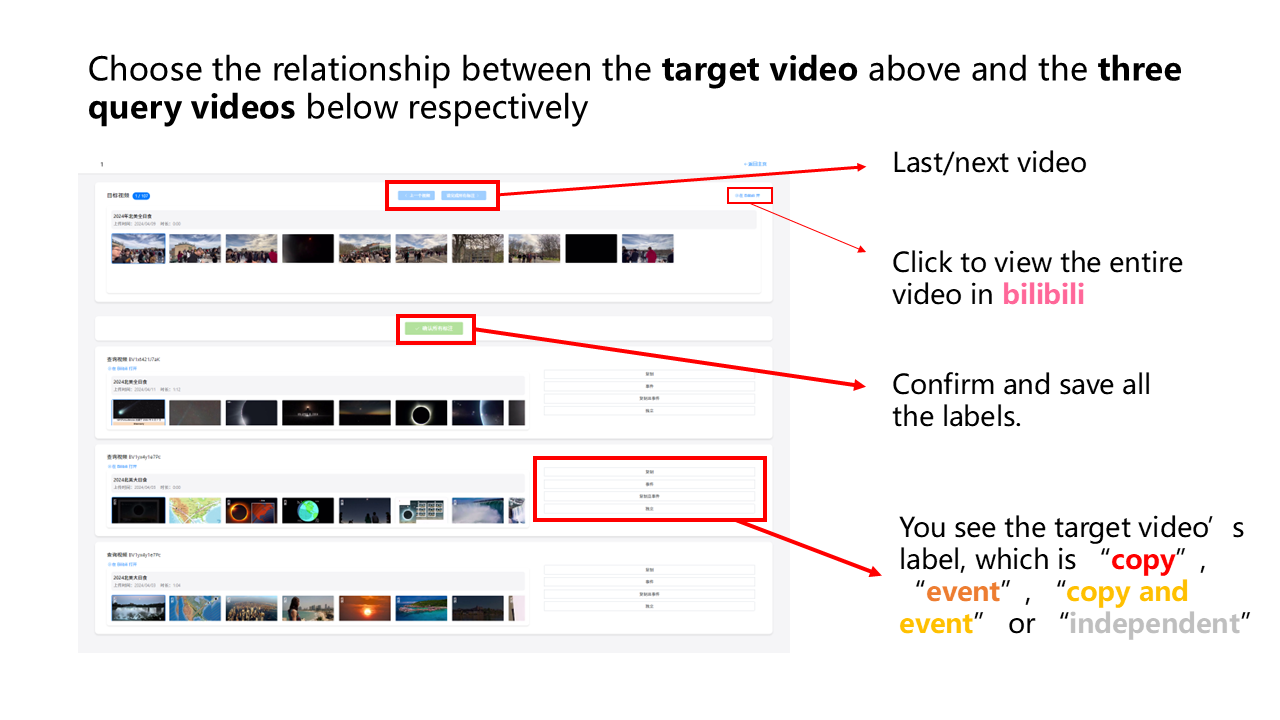}}%
    \hspace{0.0\textwidth}%
    \fbox{\includegraphics[width=0.49\textwidth]{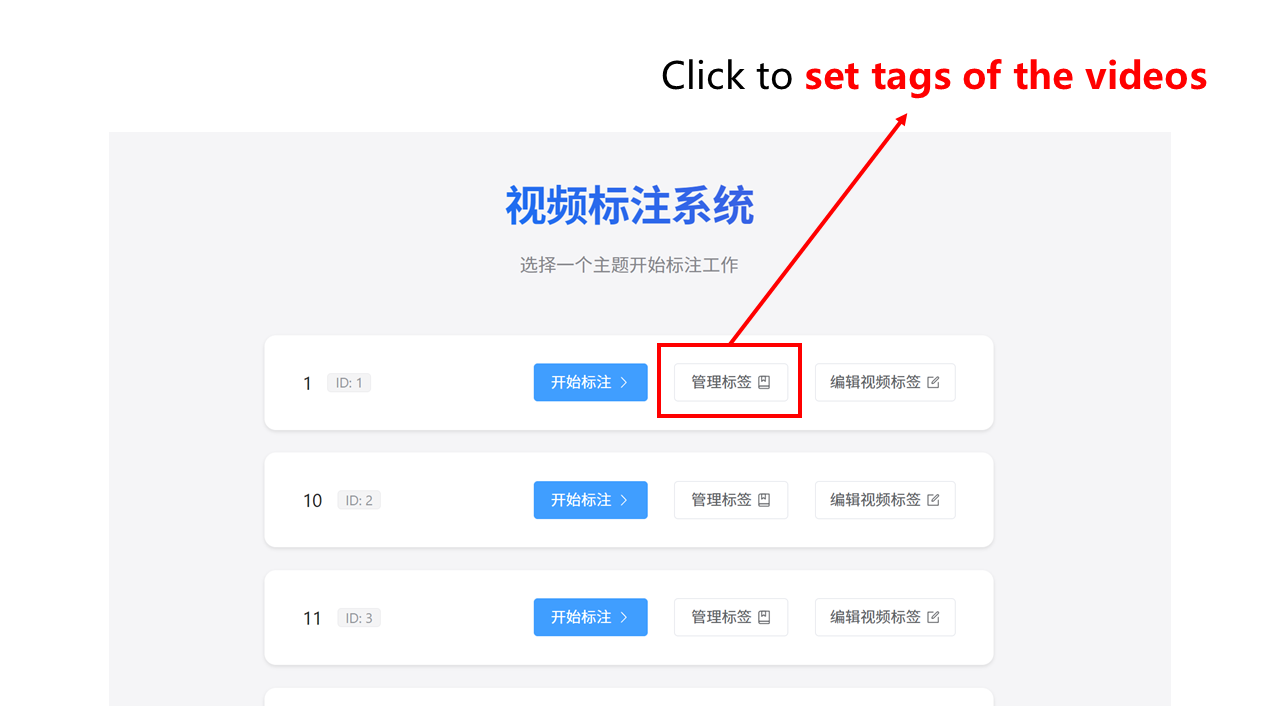}}%
    
    % 第四行
    \vspace{-0.1em}%
    \fbox{\includegraphics[width=0.49\textwidth]{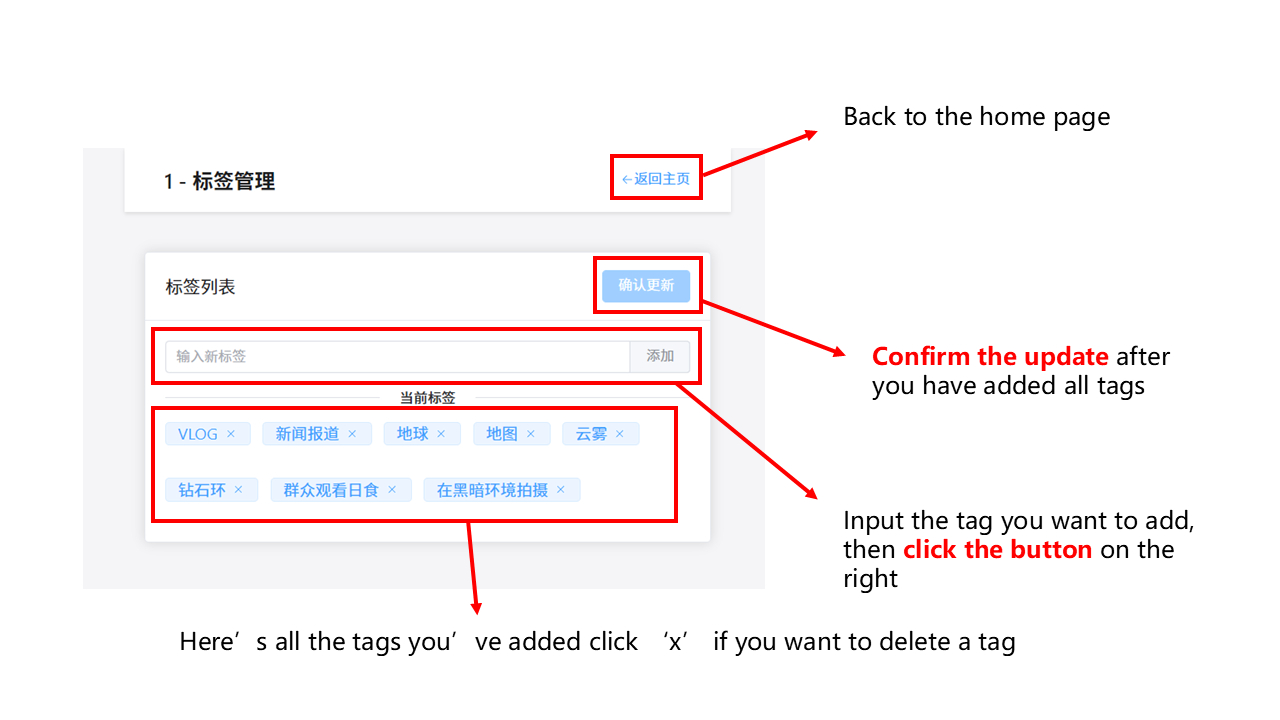}}%
    \hspace{0.0\textwidth}%
    \fbox{\includegraphics[width=0.49\textwidth]{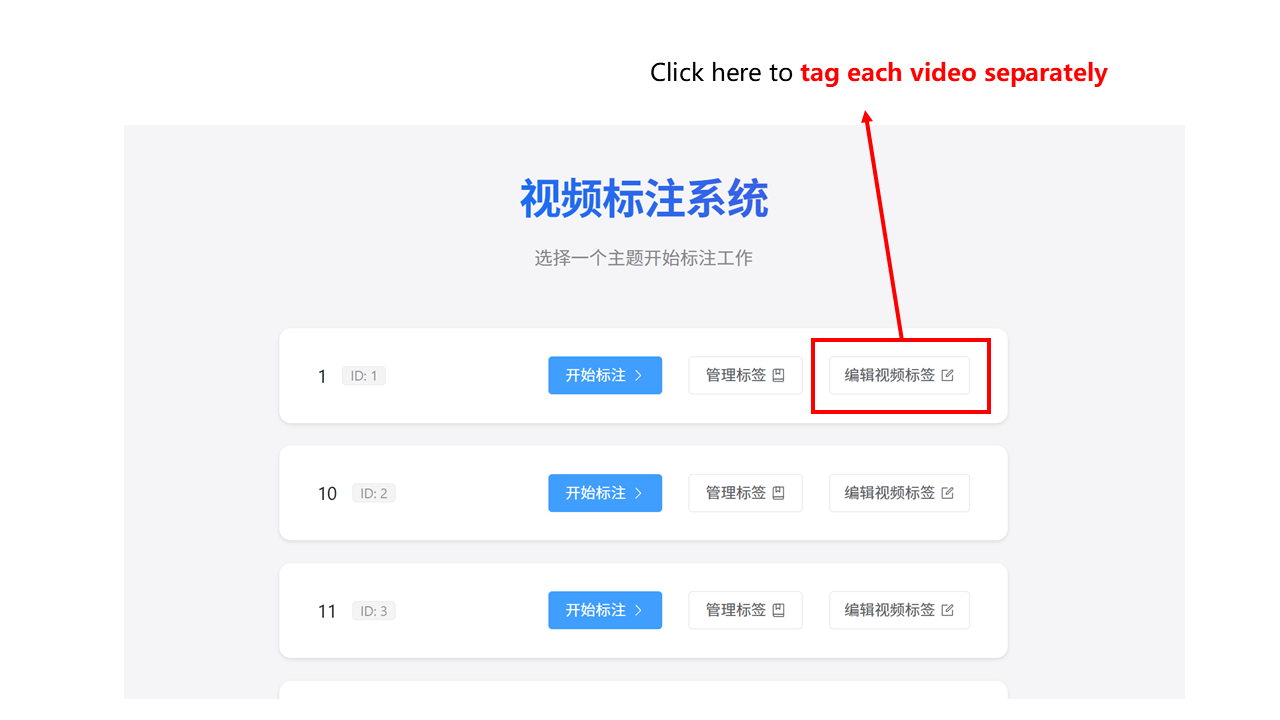}}%
    
    % 第五行（最后一张单独居中）
    \vspace{-0.1em}%
    \fbox{\includegraphics[width=0.49\textwidth]{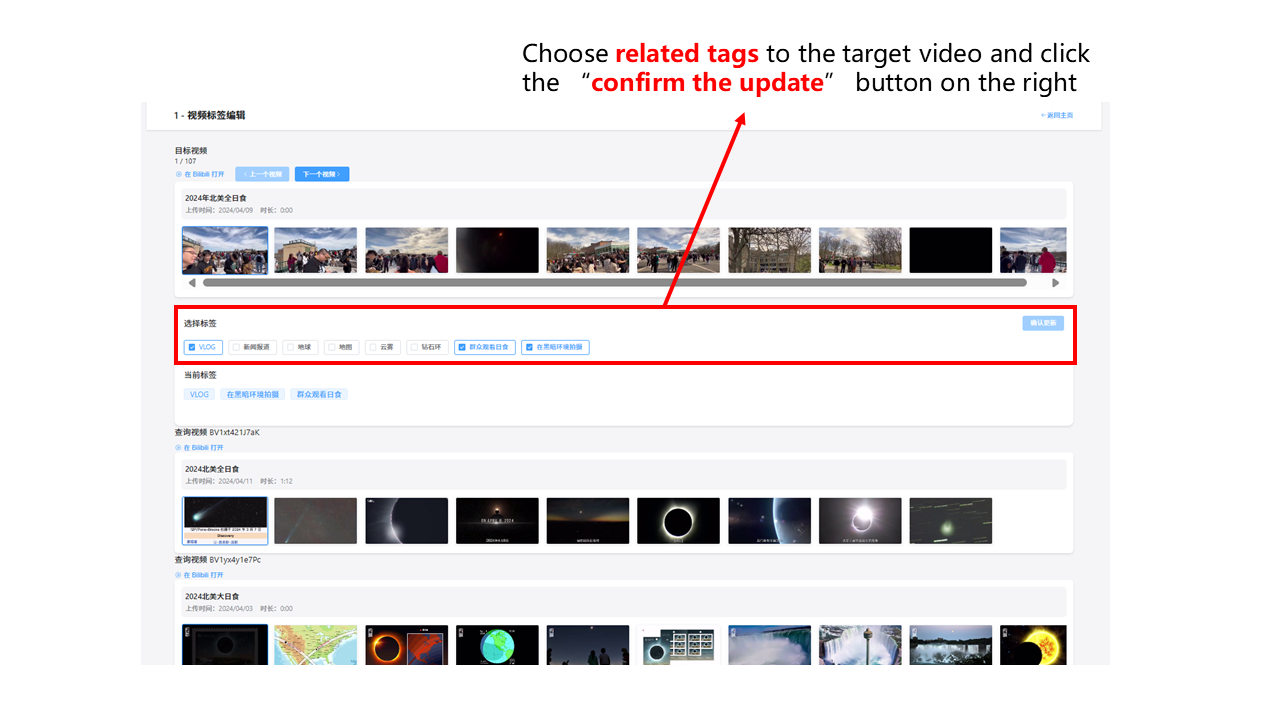}}%
\end{figure}

\section{Visualization}
\label{sec5}
\begin{figure*}[!t]
    \centering
    \includegraphics[width=1\linewidth]{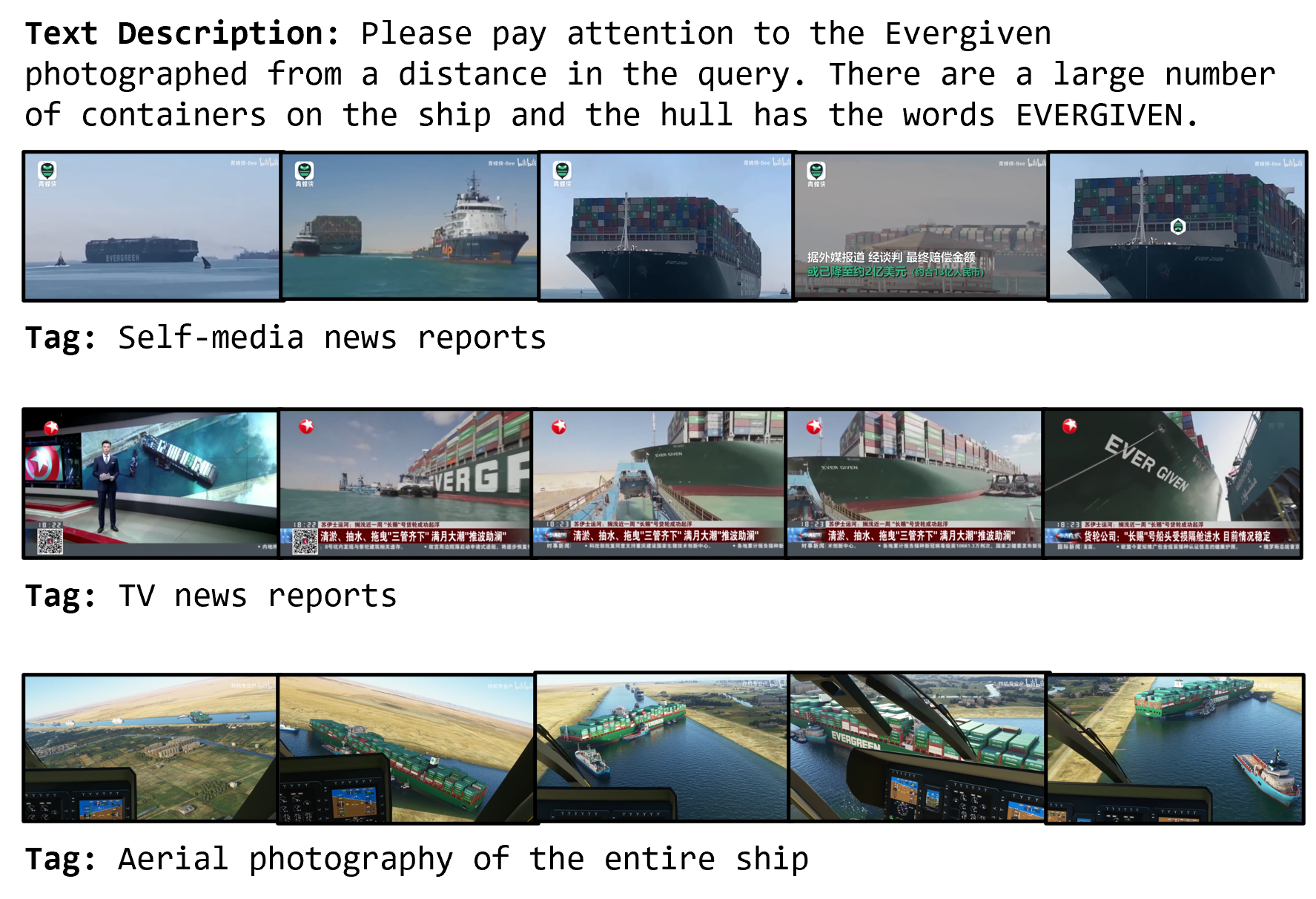}
    \caption{Visualization of three relevant videos on the News partition.}
    \label{fig: news}
\end{figure*}

\begin{figure*}[!t]
    \centering
    \includegraphics[width=1\linewidth]{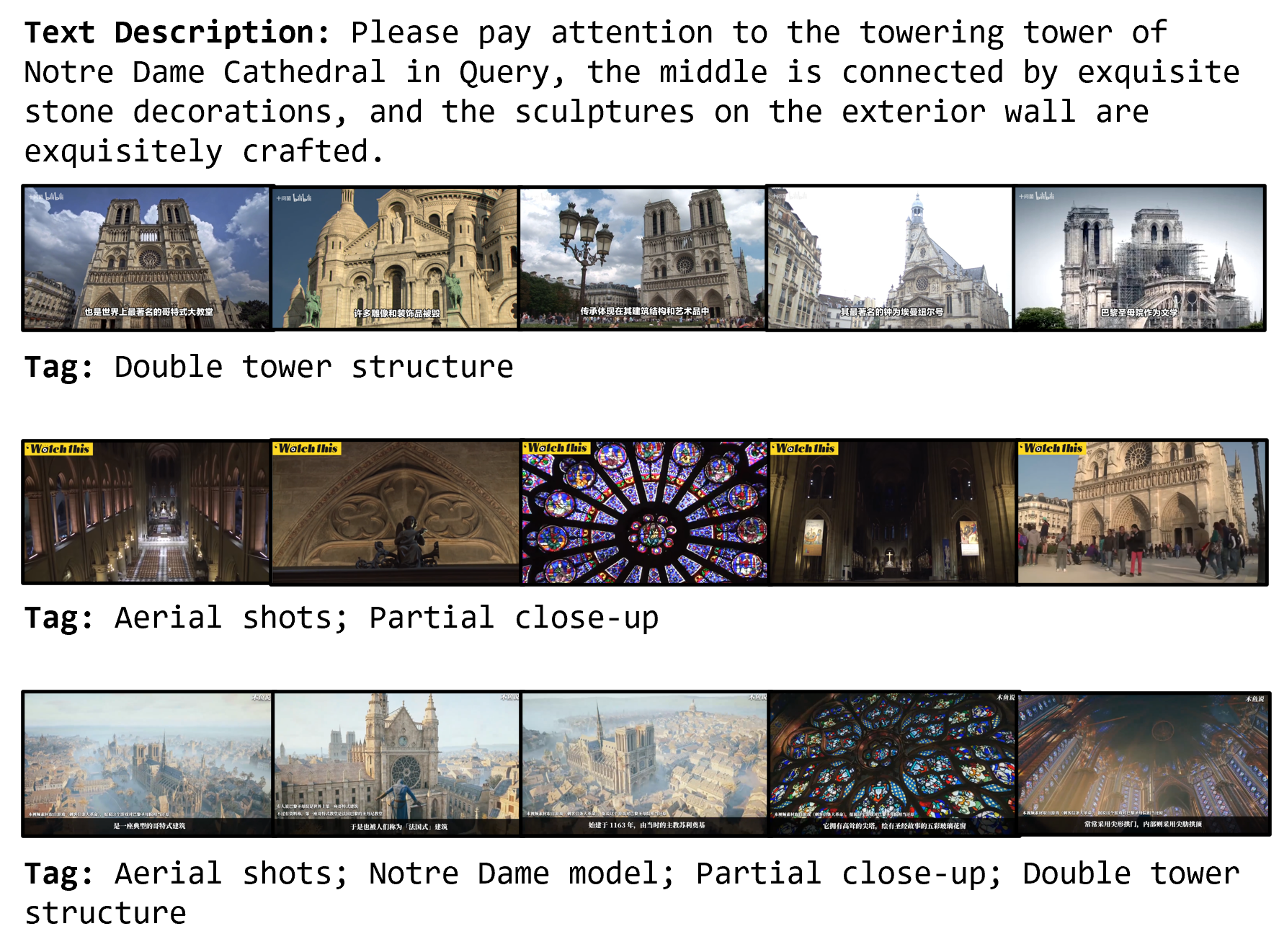}
    \caption{Visualization of three relevant videos on the Region partition. The third video comes from a computer game and brings more challenges.}
    \label{fig: region}
\end{figure*}

\begin{figure*}[!t]
    \centering
    \includegraphics[width=1\linewidth]{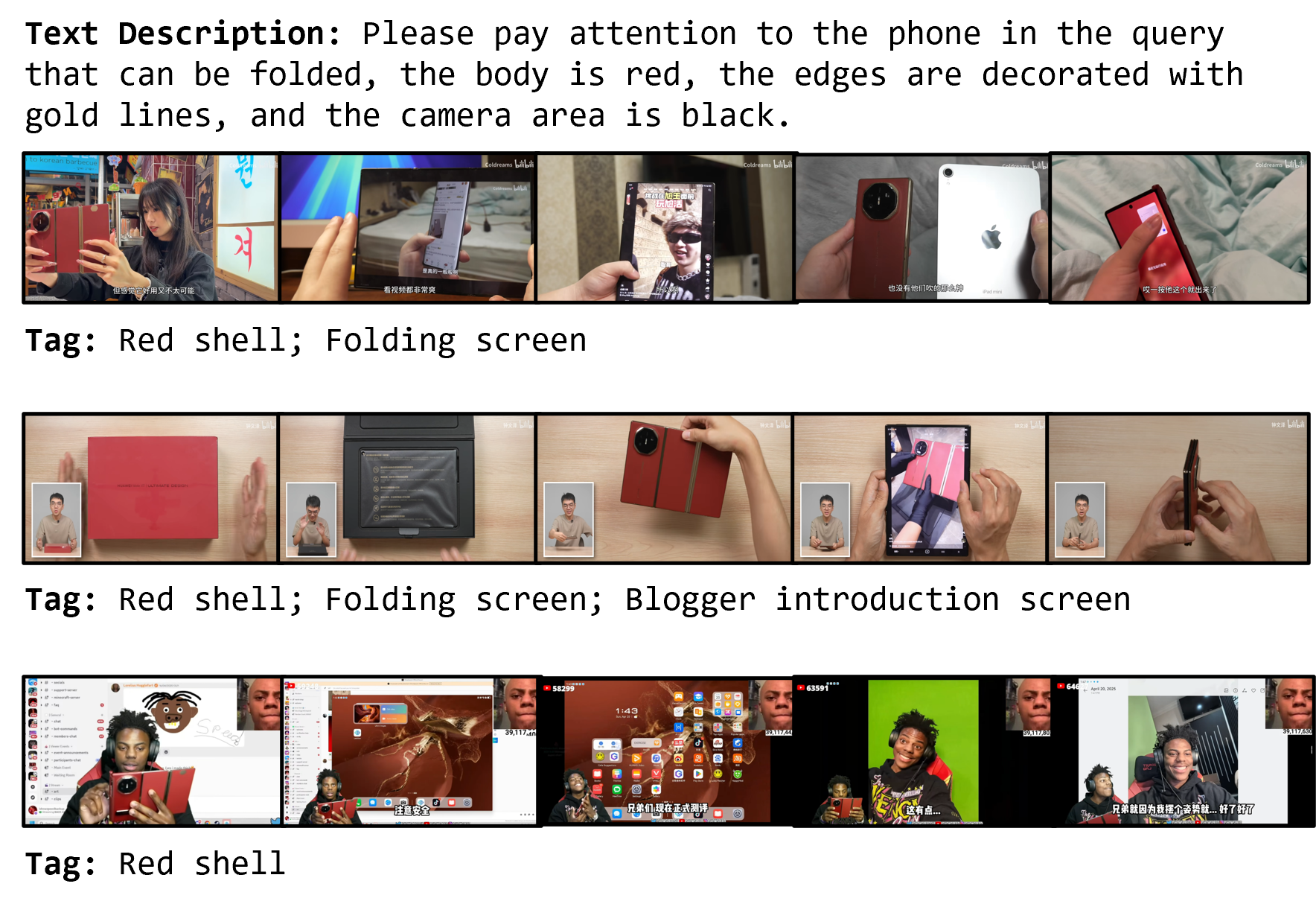}
    \caption{Visualization of three relevant videos on the Instance partition. The different forms of mobile phones and their small proportion on the screen pose challenges.}
    \label{fig: instance}
\end{figure*}

\begin{figure*}[!t]
    \centering
    \includegraphics[width=1\linewidth]{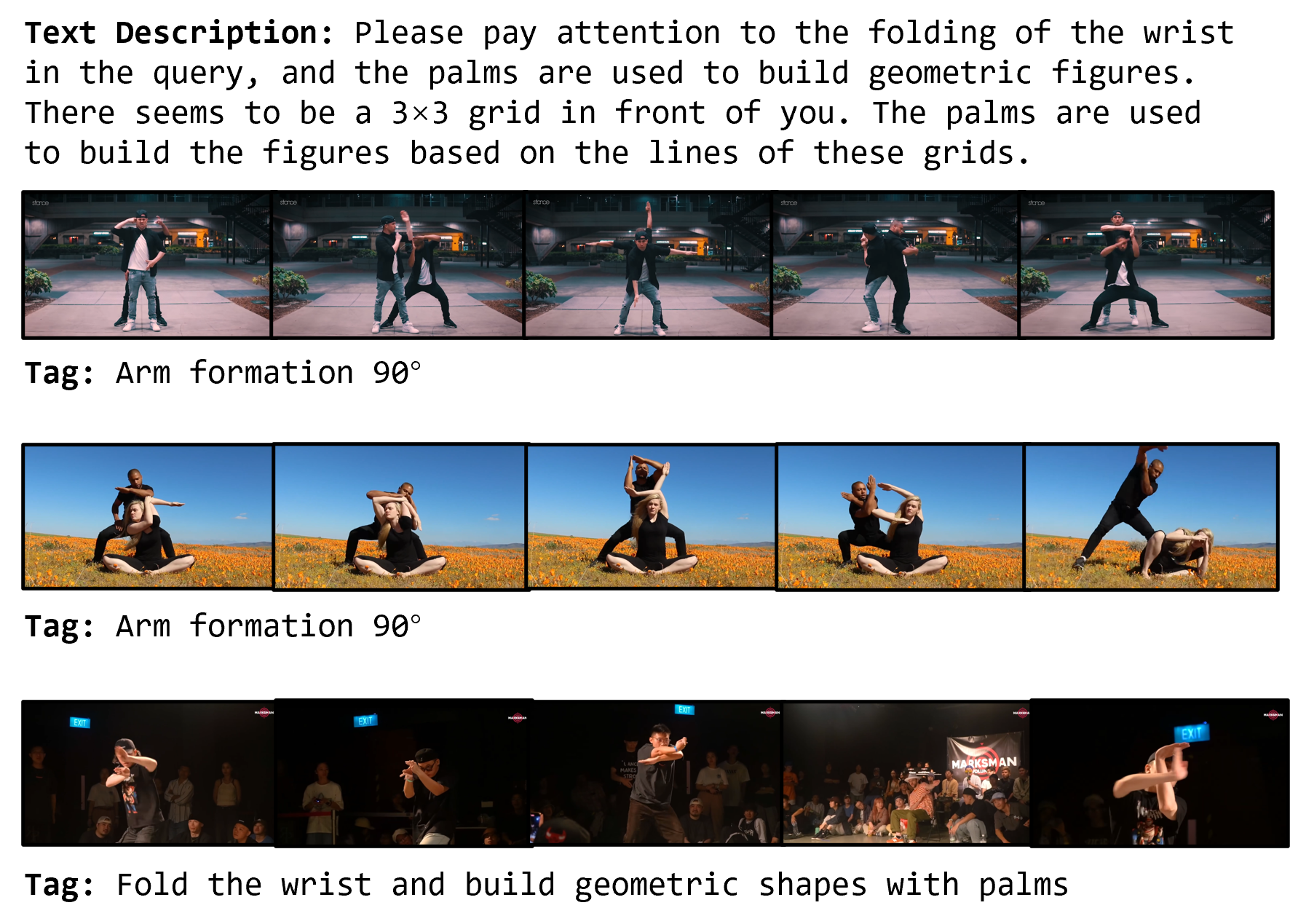}
    \caption{Visualization of three relevant videos on the Dance partition. The interference of background, characters, and the number of people poses a huge challenge to action-level retrieval.}
    \label{fig: dance}
\end{figure*}

\begin{figure*}[!t]
    \centering
    \includegraphics[width=1\linewidth]{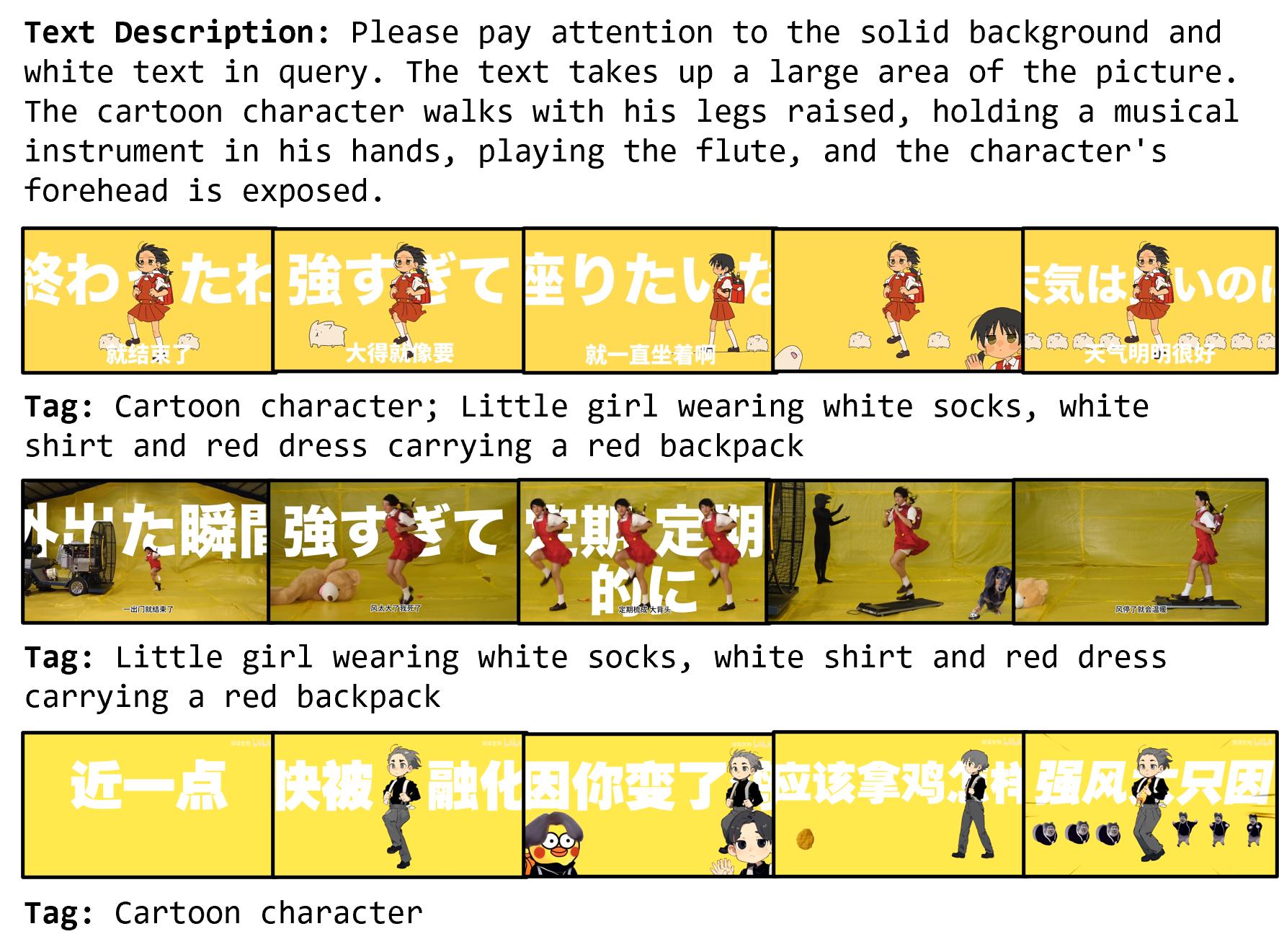}
    \caption{Visualization of three relevant videos on the Others partition. This type of video is created based on common popular elements and video styles, with rich semantic information.}
    \label{fig: others}
\end{figure*}

Figure \ref{fig: news}, \ref{fig: region}, \ref{fig: instance}, \ref{fig: dance} and \ref{fig: others} provide several relevant examples of different partitions from MUVR, with a text description of the query video and the tag of each video.


\end{document}